\documentclass[accepted]{uai2026}

\usepackage[american]{babel}

\usepackage{mathtools}  
\usepackage{amssymb}    
\usepackage{amsthm}     
\usepackage{siunitx}
\usepackage{natbib}
\usepackage{booktabs}   
\usepackage{graphicx}   
\usepackage{subcaption} 
\usepackage{multirow}   
\usepackage{tabularx}
\usepackage{float}      
\usepackage{hyperref}
\hypersetup{
 colorlinks=true,
 linkcolor=blue,
 urlcolor=blue,
 citecolor=blue
}
\usepackage[table]{xcolor} 
\usepackage{tikz}
\usepackage{pgfplots}
\pgfplotsset{compat=1.18}
\usepgfplotslibrary{groupplots}

\usepackage{algorithm}
\usepackage{algorithmic}

\usepackage{microtype}
\usepackage{hyperref}
\hypersetup{
 colorlinks=true,
 linkcolor=blue,
 urlcolor=blue,
 citecolor=blue
}
\usepackage[normalem]{ulem} 
\usepackage{listings}

\usepackage{cuted}
\usepackage{mathtools}           
\usepackage{amssymb}             
\usepackage{amsthm} 
\theoremstyle{plain}

\theoremstyle{definition}

\theoremstyle{remark}

\newenvironment{highlight}{}{}
\newcommand{\update}[1]{#1}

\definecolor{group1}{HTML}{F2F2F2} 
\definecolor{group2}{HTML}{FFFFFF} 
\definecolor{group3}{HTML}{F9F9F9}
\definecolor{group4}{HTML}{FFFFFF} 
\definecolor{best}{HTML}{D7191C} 
\definecolor{second}{HTML}{008080}

\title{Bandwidth Selection in Kernel Density Estimation for Model Calibration}

\author[1]{\href{han.zhou@kuleuven.be}{Han~Zhou}{}}
\author[1]{Teodora Popordanoska}
\author[1]{Matthew~B.~Blaschko}
\affil[1]{%
    Department ESAT, Center for Processing Speech and Images\\
    KU Leuven\\
    Leuven, Belgium
}

\begin{document}
\maketitle
\begin{abstract}
As deep learning models are increasingly deployed in high-stakes applications, providing well-calibrated uncertainty estimates has become as critical as achieving high predictive accuracy. While Kernel Density Estimation (KDE) has emerged as a smooth and continuous alternative to traditional binning for quantifying miscalibration, its reliability is heavily dependent on the choice of the kernel bandwidth. Standard selection techniques, such as Maximum Likelihood Estimation (MLE), often fail to produce optimal bandwidths for calibration tasks. In this work, we introduce Risk Alignment (RA), a novel optimization framework that determines the optimal bandwidth by aligning KDE-reconstructed risk with empirical risk. We theoretically demonstrate that this alignment minimizes calibration estimation bias across the data distribution, establishing a principled bandwidth selection criterion applicable to various metrics, including the challenging case of canonical calibration error. Extensive experiments across multiple architectures and datasets show that RA consistently outperforms standard bandwidth selection methods, yielding more reliable calibration assessments. \update{Our implementation is available at~\url{https://github.com/han678/KDE-Bandwidth-Learning}.}
\end{abstract}

\section{Introduction}\label{sec:intro}
Over the past decade, deep learning has propelled neural networks to achieve, and occasionally exceed, human-level proficiency in domains ranging from image recognition~\citep{haggenmuller2021skin} to natural language understanding~\citep{achiam2023gpt, liu2024deepseek}. Despite these gains in predictive power, high accuracy does not inherently guarantee that a model’s confidence scores are well-aligned with the true likelihood of its outcomes. This discrepancy, termed miscalibration~\citep{guo2017calibration}, poses significant risks in high-stakes environments where reliable uncertainty quantification is as critical as the prediction.

To evaluate this alignment, various metrics have been proposed to quantify calibration error~\citep{naeini2015obtaining, guo2017calibration, ding2020revisiting, popordanoska2022consistent}. However, the inherent challenge in evaluating these metrics stems from our lack of access to the true conditional probability $R = \mathbb{E}[y \mid f(x)]$, which necessitates reliance on empirical approximations. The most prevalent approach, the binning method~\citep{guo2017calibration}, partitions the unit interval into discrete segments to approximate the conditional probability. While straightforward, binning introduces artificial discontinuities and suffers from high variance, fundamentally limiting its estimation performance. Specifically, these estimators are often asymptotically inconsistent~\citep{widmann2019calibration} and highly sensitive to the partitioning scheme~\citep{nixon2019measuring}. Moreover, they suffer from the curse of dimensionality in multi-class settings, where exponential bin growth leads to severe data sparsity~\citep{ding2020revisiting, arrieta2022metrics}. Recognizing these limitations, another line of research leverages Kernel Density Estimation (KDE) to provide a differentiable alternative~\citep{popordanoska2022consistent}. By employing a kernel function to aggregate information across the probability space, this approach yields a smooth representation of the conditional probability $R$, thereby circumventing the information loss inherent in discretization.

However, the reliability of these KDE-based estimators remains highly contingent on a critical tuning parameter: the kernel bandwidth~\citep{buch2005kernel, silverman2018density, blasioksmooth, gruber2025optimizing}. While standard MLE methods~\citep{duin1976choice} are optimal for general density estimation, they do not explicitly account for the estimation variance inherent in calibration metrics. In practice, the likelihood-based objective often favors narrow bandwidths that overfit to local sampling fluctuations rather than capturing the underlying calibration profile. This mismatch leads to noisy estimates that limit the reliability of the resulting calibration surface. In this work, we tackle the bandwidth selection challenge by proposing the Risk Alignment objective (RA) framework. Unlike density-centric methods, RA utilizes a principled optimization objective grounded in the decomposition of proper scoring rules to determine an optimal, variance-robust bandwidth. This objective is applicable across all notions of calibration metrics, including the notoriously difficult canonical calibration. Our contributions are as follows:
\begin{itemize} 
\setlength\parskip{0pt}  
\item We introduce a novel bandwidth selection framework for KDE-based calibration estimators in Section~\ref{sec:ra}, formulated as a risk estimation alignment (RA) optimization objective. 
\item  We provide a theoretical analysis of the bias-variance trade-off in calibration estimation, demonstrating that the RA objective recovers the underlying conditional probability by minimizing aggregate bias. 

\item We empirically demonstrate that our method consistently selects superior bandwidths compared to MLE, leading to better calibration estimation. 
\end{itemize}

\section{Related work}
In this section, we provide an overview of existing literature on quantifying model calibration errors.

\textbf{Binning-Based Estimation.} A prominent line of work involves binning-based methods, such as the widely used Expected Calibration Error (ECE)~\citep{guo2017calibration}, which estimates $R$ by partitioning the prediction space into discrete intervals and averaging labels within each bin. Early techniques utilized equal-width or adaptive binning~\citep{ding2020revisiting}, but these suffer from sensitivity to the binning scheme~\citep{kumar2019verified}. However, a fundamental limitation persists: these approaches treat $R$ as a \textit{piecewise constant} function, which introduces notable statistical bias~\citep{roelofs2022mitigating} and renders the metric non-differentiable. This lack of smoothness precludes their integration into gradient-based optimization frameworks, preventing their utility as differentiable calibration regularizers during model training.

\textbf{Kernel Density Estimation (KDE).}  Another line of work utilizes KDE to circumvent the limitations of discrete binning by providing smooth, differentiable approximations of conditional probability. While early approaches~\citep{zhang2020mix} faced scalability issues in multi-class settings, recent methods have introduced specific kernel choices, such as RBF~\citep{blasioksmooth} and Dirichlet kernel~\citep{popordanoska2022consistent, popordanoska2024consistent}, to estimate calibration error. However, these estimators remain highly sensitive to kernel bandwidth. Current practices often rely on maximum likelihood estimation (MLE) or manual visual inspection. This often leads to suboptimal bandwidth values that fail to generalize, causing the estimated conditional probability $R$ to overfit to local sampling noise rather than capturing the true underlying probability across the data distribution. A recent advancement~\citep{gruber2025optimizing} introduces an MSE-based risk framework to optimize calibration estimators by reformulating the task as a regression problem. However, this approach requires expensive cross-validation for its regularization parameter and entails $\mathcal{O}(n^3)$ complexity due to eigenvalue decompositions—limiting its scalability compared to the $\mathcal{O}(n^2)$ efficiency of KDE method~\citep{popordanoska2024consistent}. To address this, RA retains the KDE framework but derives a direct optimization objective from the bregman decomposition of the proper scoring rules. This approach remains naturally robust to sampling noise, achieving $\mathcal{O}(n^2)$ efficiency while bypassing the expensive cross-validation required by prior methods.

\section{Notation and Preliminaries}

\textbf{Multiclass Classification.}
Consider a multiclass classification task with $k$ classes. We aim to learn a classifier $f: \mathcal{X} \to \Delta^{K-1}$ that maps an input $x$ from the feature space $\mathcal{X} \subseteq \mathbb{R}^d$ to the probability simplex $\Delta^{K-1} = \{ f^{(k)}(x) \in [0,1] :\sum_{i=1}^k f^{(k)}(x_i) = 1 \}$. The output vector $f(x)$ represents the predicted class probabilities. Let $y \in \{0,1\}^k$ be the one-hot encoded label, where $(x, y)$ follows a joint distribution $\mathbb{P}(x, y)$.

The performance of a classifier is evaluated by the expected risk:
\begin{equation}
    \mathcal{R}(f) = \mathbb{E}_{(x,y) \sim \mathbb{P}(x, y)}[\ell(f(x), y)],
\end{equation}
where $\ell$ is a loss function. A loss $\ell$ is called \textbf{proper} if the risk is minimized by the Bayes classifier $f^*(x) \coloneqq \mathbb{E}[y|x]$~\citep{gneiting2007strictly}. 

\textbf{Bregman Divergence.}
Every differentiable proper loss is uniquely associated with a Bregman divergence~\citep{Bregman1967}. Given a continuously differentiable and strictly convex function $F: \Delta^{K-1} \to \mathbb{R}$, the associated Bregman divergence is defined as:
\begin{equation}
    D_F(p, q) \coloneqq F(p) - F(q) - \langle \nabla F(q), p - q \rangle
\end{equation}
which measures the discrepancy between the value of $F$ at $p$ and its linear approximation anchored at $q$. Notable special cases include the squared Euclidean distance $D_F(p, q) = \|p - q\|_2^2$ induced by the $L_2$ norm $F(p) = \|p\|_2^2$ and the Kullback–Leibler (KL) divergence $D_F(p, q) = \mathrm{KL}(p \| q)$ induced by the negative Shannon entropy $F(p) = \sum_{i=1}^k p_i \ln p_i$.

\textbf{Bregman Decomposition.}
A well-known result~\citep{Broecker2008,kull2015novel,ovcharov2015existence, popordanoska2022consistent,NEURIPS2022_3915a87d, berta2025rethinking} in the study of proper scoring rules is that the risk of a proper loss can be decomposed into calibration error and refinement. Let $R = \mathbb{E}[y | f(x)]$ denote the true conditional probability, the risk $\mathcal{R}(f)$ is decomposed as follows:
\begin{equation}\label{eq:riskdecompose} 
    \mathcal{R}(f) = \underbrace{\mathbb{E} \left[ D_F \bigl( R, f(x) \bigr) \right]}_{\text{CE}_F(f) \text{ (Proper Calibration Error)}} + \underbrace{\mathbb{E} \left[F(y) -F \bigl( R \bigr) \right]}_{\text{REF}_F(f) \text{ (Refinement)}}
\end{equation}where the calibration error measures how well the predicted probabilities match the true conditional probabilities; the refinement reflects the model’s ability to \emph{distinguish} samples. We note that the Generalized Entropy term $\mathbb{E}[F(Y)]$~\citep{gneiting2007strictly} is independent of the predictor $f$, so it often excluded for optimization and model
comparison. In this framework, the proper calibration error ($\text{CE}_F$) can be related to the specific strictly convex function $F$ that defines the corresponding Bregman divergence. One special case is the $L_2$ calibration error~\citep{murphy1973new}, defined as:
\begin{equation}
\mathrm{CE}_2^2(f)=\mathbb{E}\left[\|R -f(x)\|_2^2\right].
\end{equation}

While canonical calibration assesses the model's overall reliability, it is often desirable to evaluate calibration at a per-class level, particularly in the presence of class imbalance. \update{Under the $L_2$ loss, the class-wise calibration error is defined as the sum of squared binary calibration errors: 
\begin{align} \label{eq:classwise_l2_ce}
\text{CWCE}_2^2(f)= \sum_{k=1}^K \mathbb{E} \left[ \bigl(R^{(k)} - f^{(k)}(x)\bigr)^2 \right].
\end{align} where $R^{(k)}$ and $f^{(k)}(x)$ denote the true and predicted probabilities for class $k$, respectively.} This multi-class evaluation~\citep{nixon2019measuring,panchenko2022class,popordanoska2024consistent} can be done by decomposing the problem into $K$ independent binary tasks via a one-versus-rest (OvR) approach. Under this decomposition, the expected class-wise risk is defined as the sum of these binary risks. Consequently, for any proper scoring rule induced by $F$, the class-wise risk admits an individual Bregman decomposition for each class, the details of which are provided in Section~\ref{sec:classwiseOvR}.

\section{Kernel density based calibration error estimator}
\subsection{Bregman Decomposition}
Based on the risk decomposition in Eq.~\eqref{eq:riskdecompose}, we can derive empirical estimators for calibration and refinement. Given a dataset $\mathcal{D} = \{x_i, y_i\}_{i=1}^n$ and assuming that the true conditional probability $R$ is known, then it leads to the the Bregman formulation of calibration error estimator~\citep{gruber2022better,popordanoska2024consistent}:
\begin{align}  \label{eq:ce_bregman_def} 
\widehat{\text{CE}}_{F}(f) \approx \frac{1}{n} \sum_{i=1}^{n} \Bigl( &F(R_i) - F(f(x_i)) \nonumber \\
&- \langle \nabla F(f(x_i)), R_i - f(x_i) \rangle \Bigr)
\end{align}
where $R_i$ is the true conditional probability at $f(x_i)$. Similarly, the refinement estimator is given by
\begin{equation} ~\label{eq:ref_bregman_def}
\widehat{\mathrm{REF}}_F(f)
=\frac{1}{n} \sum_{i=1}^n \bigl(F(y_i) - F(R_i)\bigr).
\end{equation}
In particular, under the squared $L_2$ loss, the resulting calibration error estimator reduces to
\begin{equation}
        \widehat{\text{CE}}_2^2(f) = \frac{1}{n} \sum_{i=1}^n \left\| R_i - f(x_i) \right\|_2^2 
\end{equation}
while for KL divergence, it takes the form
\begin{equation}
    \widehat{\text{CE}}_{\text{KL}}(f) =\frac{1}{n} \sum_{i=1}^n\left\langle\ R_i, \log \left( \frac{ R_i}{f(x_i)}\right)\right\rangle.
\end{equation}

\subsection{Kernel Density estimator for $R$}
In practice, the true conditional expectation $R$ is unknown and must be inferred from observed data. The fidelity of any calibration error estimator—whether based on discrete binning or kernel-based methods—fundamentally relies on the accurate recovery of the conditional label distribution $R$, as the discrepancy between $R$ and the predicted confidence $f(x)$ defines the calibration error. \citet{popordanoska2022consistent} propose a KDE approach that estimates $R$ as a locally weighted average of observed labels, yielding a smooth, differentiable surface that reflects the underlying conditional label distribution.

\paragraph{Kernel Density Estimation.} While the Beta kernel is typically employed for binary classification, the multi-class setting—where predictions reside on the probability simplex $\Delta^{K-1}$—is naturally modeled with a density estimator based on the Dirichlet kernel, defined as~\citep{popordanoska2022consistent, popordanoska2024consistent}:
\begin{equation}
k_{\text{Dir}}(f(x_i), f(x_j); h) = \frac{\Gamma(\sum_{k=1}^K \alpha_{jk})}{\prod_{k=1}^K \Gamma(\alpha_{jk})} \prod_{k=1}^K f(x_i)_k^{\alpha_{jk}-1},
\end{equation}
where $\alpha_j = \frac{f(x_j)}{h} + 1$ and $h > 0$ is the smoothing bandwidth. To ensure a consistent and asymptotically unbiased estimation of $R_i$ at each observed point $f(x_i)$, we utilize a leave-one-out (LOO) estimator:
\begin{equation} \label{eq:estimate_R}
    \hat{R}_i = \frac{\sum_{j \neq i } k_{\operatorname{Dir}}(f(x_i), f(x_j); h) y_j}{\sum_{j \neq i}  k_{\operatorname{Dir}}(f(x_i), f(x_j); h)}.
\end{equation}
By substituting $\hat{R}_i$ for the true conditional probability $R_i$ in Eq.~\eqref{eq:ce_bregman_def} and \eqref{eq:ref_bregman_def}, we obtain a kernel density-based estimators for both calibration error and refinement. However, the fidelity of these estimators is intrinsically tied to the smoothing bandwidth $h$. This parameter dictates the balance between the local sensitivity and the global stability of the conditional probability manifold. Consequently, identifying an optimal $h$ is not merely a numerical necessity but a prerequisite for ensuring that the empirical metrics reflect the true underlying calibration.

\subsection{Bandwidth selection}
Choosing the bandwidth parameter $h$ is a long-recognized and critical step~\citep{loader1999local, popordanoska2022consistent, gruber2025optimizing}, as it governs the capacity of the selected kernel to aggregate information from neighboring samples. Improperly tuned $h$ can lead to either over-smoothing (high bias) or excessive volatility (high variance) in the conditional probability estimate.

\paragraph{Log-likelihood Maximization (MLE).} 
A standard heuristic for bandwidth selection, adopted by \citet{popordanoska2022consistent}, is the maximization of the leave-one-out (LOO) log-likelihood. This nonparametric method focuses on the generative probability of the features in the prediction space.
Given bandwidth $h$, the LOO density estimate at $f(x_i)$ is
defined as
\begin{equation}
\widehat{p}_{-i}\!\left(f(x_i);h\right)
=
\frac{1}{n-1}\sum_{j\neq i} k_{\operatorname{Dir}}\!\left(f(x_i), f(x_j); h\right).
\end{equation}
 The optimal bandwidth $h^\star$ is then sought by maximizing the aggregate log-likelihood:
\begin{equation}
\mathcal{L}_{\text{MLE}}(h) = \sum_{i=1}^n
\log \widehat{p}_{-i}\!\left(f(x_i);h\right).
\end{equation}
While MLE is effective for general density estimation, it fails in calibration due to a critical loss of generality. As the sample size $n$ increases, the MLE objective favors pathologically small bandwidths (See Fig.~\ref{fig:mle_bw}). By pursuing local density maximization, MLE selects infinitesimal $h$ that causes the kernel to contract around individual data points, which often leads to overfitting. Consequently, the estimator $\hat{R}_i$ degenerates from the true conditional probability into a nearest-neighbor label sampler. Instead of recovering a smooth, generalizable reliability curve, it produces sharp spikes that capture stochastic sampling noise. As derived in Section~\ref{sec:bias-variance-trap}, this overfitting creates a "variance-bias trap": the excessive estimation variance is misattributed to the calibration error estimator, yielding a persistent positive bias.

\section{Calibration-Oriented Bandwidth Selection} \label{sec：RA}
We revisit bandwidth optimization through a calibration-oriented lens by utilizing the decomposition of proper scoring rules. We show that the estimated risk bias is  aligned with the conditional probability bias through a scaling factor  related to the predicted probability. This insight allows us to develop Risk Alignment (RA), a framework that identifies the optimal bandwidth $h$ by ensuring consistency between the KDE-reconstructed risk and the empirical risk observed from samples.

\subsection{Risk Alignment} \label{sec:ra}
For the binary case under the $L_2$ risk (Brier Score) with predicted confidence $f(x) \in [0, 1]$ and label $y \in \{0, 1\}$, let $R = \mathbb{E}[y|f(x)]$ denote the true conditional probability. The expected $L_2$ risk has the form:
\begin{align} 
\mathcal{R}_{L_2}(f) = \underbrace{\mathbb{E}[(R - f(x))^2]}_{\text{CE}_2^2(f)} + \underbrace{\mathbb{E}[R(1 - R)]}_{\text{REF}_2(f)}. \label{eq:l2riskdecomposeexpectation}
\end{align}
In practice, as $R$ is inaccessible, we substitute it with the LOO-KDE estimator $\hat{R}_i$ from Eq.~\eqref{eq:estimate_R}. For a given $h$, we define the \textit{pointwise KDE-reconstructed risk} as:
\begin{align}
    \hat{r}_{\text{kde},i}(h) = \widehat{\text{ce}}_i(h) + \widehat{\text{ref}}_i(h),
\end{align}
where
\begin{align}
\widehat{\text{ce}}_i(h) = (\hat{R}_i - f(x_i))^2, \quad \widehat{\text{ref}}_i(h) = \hat{R}_i(1 - \hat{R}_i) \nonumber
\end{align}
denote the sample-wise calibration and refinement estimators, respectively. Correspondingly, let $r_i = (y_i - f(x_i))^2$ be the observed risk at sample $i$. We use lowercase notation to denote pointwise  quantities.

\begin{algorithm}[htbp]
\caption{Risk-based Bandwidth Selection}
\label{alg:grid_search}
\begin{algorithmic}[1]
\STATE \textbf{Input:} Dataset \( \mathcal{D} = \{(f_i, y_i)\}_{i=1}^n \), grid of candidates $\mathcal{H} = \{h_1, h_2, \dots, h_m\}$
\STATE \textbf{Output:} Optimal bandwidth \( h^* \)
\STATE Compute the point-wise risk $r_i$
\FOR{each candidate $h \in \mathcal{H}$}
    \STATE Compute the LOO estimates $\widehat{R}_i$ using Eq.~\eqref{eq:estimate_R}
    \STATE Evaluate $\mathcal{L}_{\text{RA}}(h) $  in Eq.~\eqref{eq:risk_opt_obj}
\ENDFOR
\STATE \textbf{Return:} $h^* = \arg\min_{h \in \mathcal{H}}\mathcal{L}_{\text{RA}}(h)$
\end{algorithmic}
\end{algorithm}

\paragraph{Risk alignment (RA) objective.} In this work, we propose an objective to recover $R$ by minimizing the aggregate discrepancy between the reconstructed and observed risks:
\begin{align} \label{eq:risk_opt_obj}
 \mathcal{L}_{\text{RA}}(h) =\sum_{i=1}^n \left| \hat{r}_{\text{kde},i}(h) - r_i \right|^p.
\end{align}
By aligning the KDE-reconstructed risk with the empirical ground truth, this framework facilitates bandwidth selection conducive to out-of-sample generalization. Unlike the MLE criterion—which is prone to the "variance-bias trap" and tends to select pathologically small bandwidths by overfitting local density fluctuations—this risk-minimization strategy favors a more conservative (larger) bandwidth. This leads to a marked reduction in estimation variance and, crucially, effectively mitigates the calibration bias induced by high-variance fluctuations (see Fig.~\ref{fig:compare_ce}). As we will demonstrate in the following analysis, by suppressing these variance-related artifacts, RA ensures a consistent estimation of $R$ across the entire prediction space.

\subsection{Decomposition of Estimation Bias} \label{sec:bias-variance-trap}

\begin{figure*}[htbp]
\footnotesize
\centering 
\begin{minipage}[t]{0.23\linewidth} 
    \centering
    \includegraphics[width=1.56in]{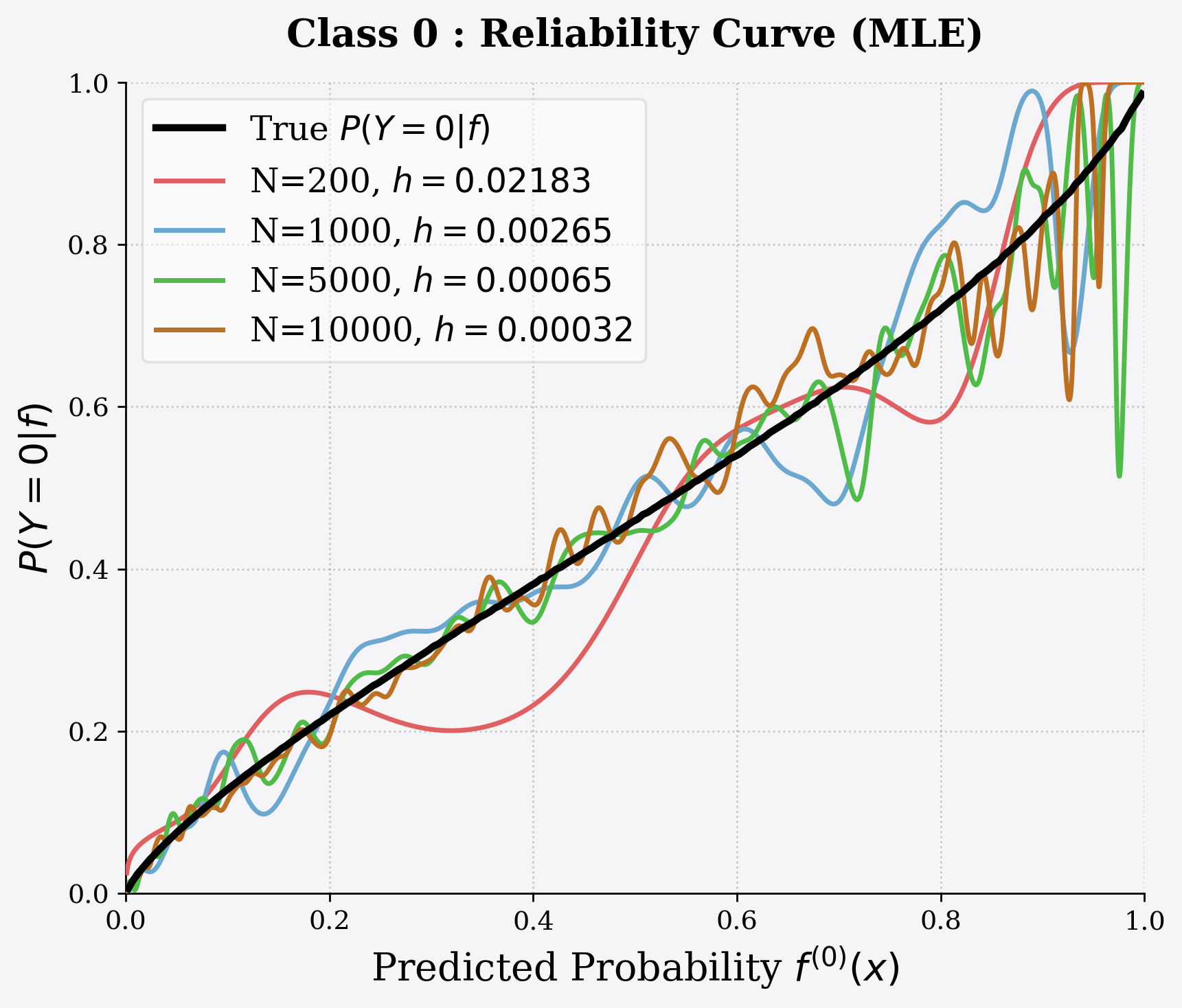}
    \subcaption{\textbf{MLE Behaviour} }
    \label{fig:mle_bw}
\end{minipage}%
\hspace{0.01\linewidth} 
\begin{minipage}[t]{0.23\linewidth} 
    \centering
    \includegraphics[width=1.56in]{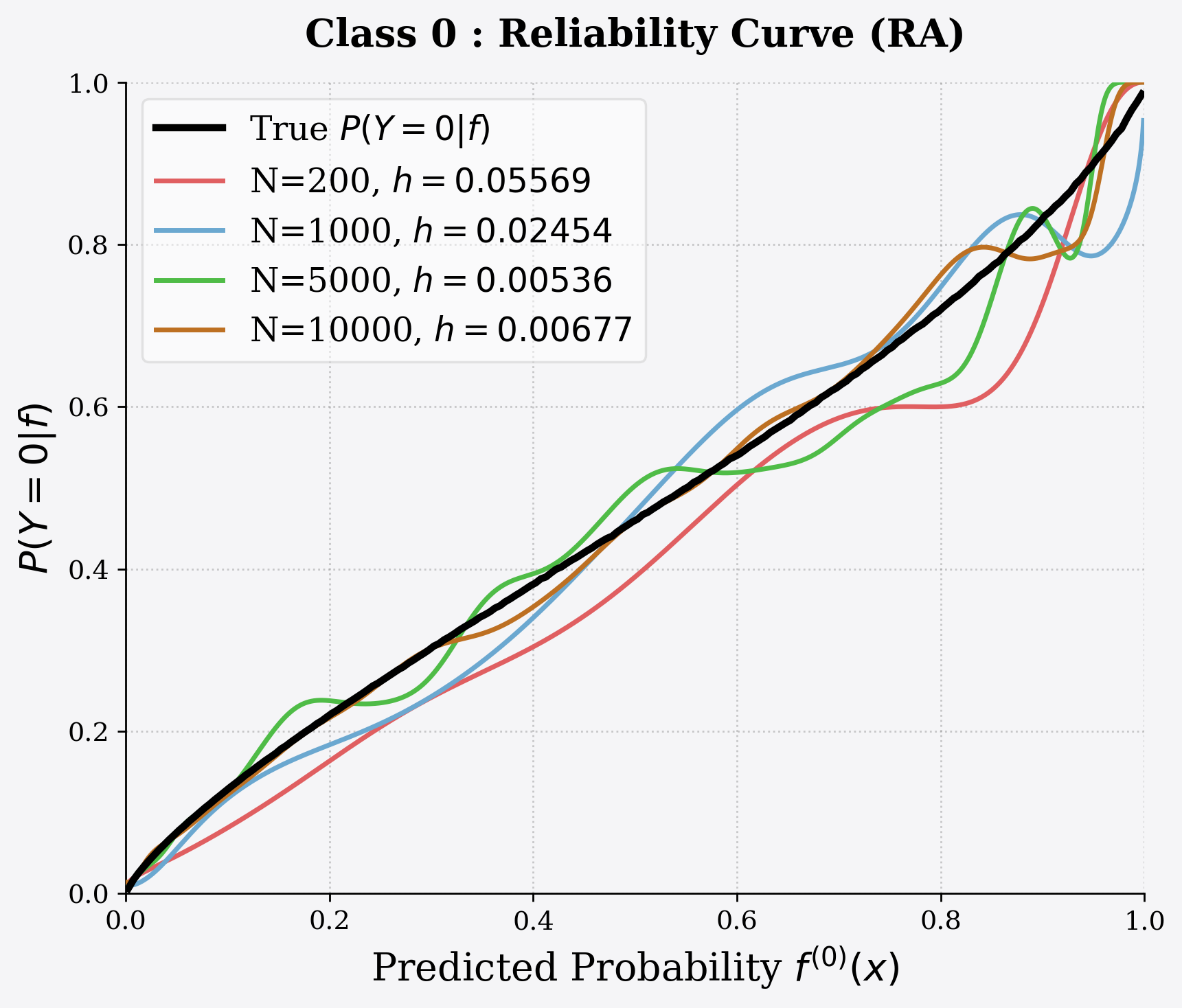}
    \subcaption{\textbf{RA Behaviour}}
    \label{fig:ra_bw}
\end{minipage}%
\hspace{0.01\linewidth} 
\begin{minipage}[t]{0.23\linewidth} 
    \centering
    \includegraphics[width=1.56in]{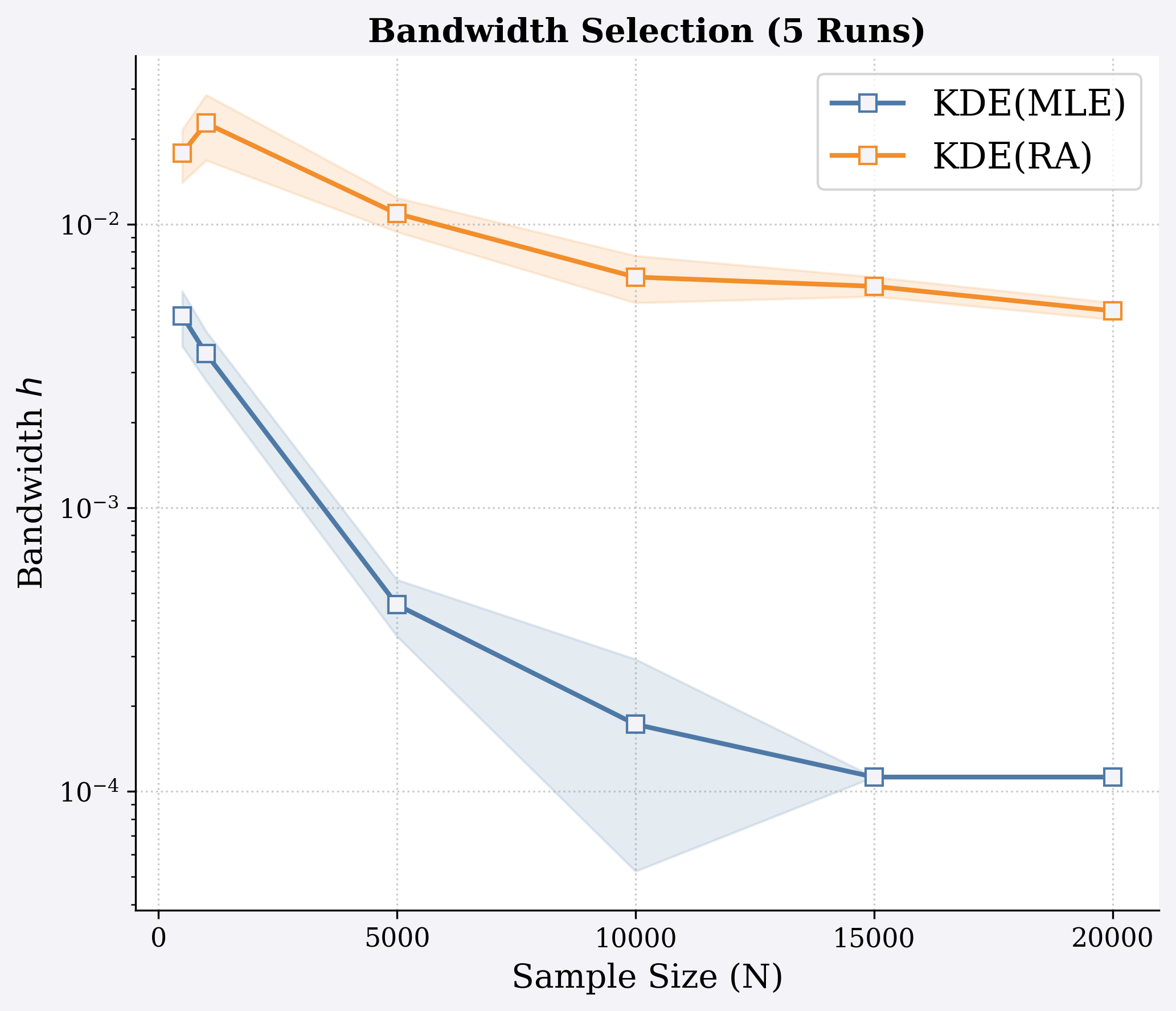}
    \subcaption{\textbf{Bandwidth}}
    \label{fig:compare_bw}
\end{minipage}%
\hspace{0.01\linewidth}  
\begin{minipage}[t]{0.23\linewidth}
    \centering
    \includegraphics[width=1.56in]{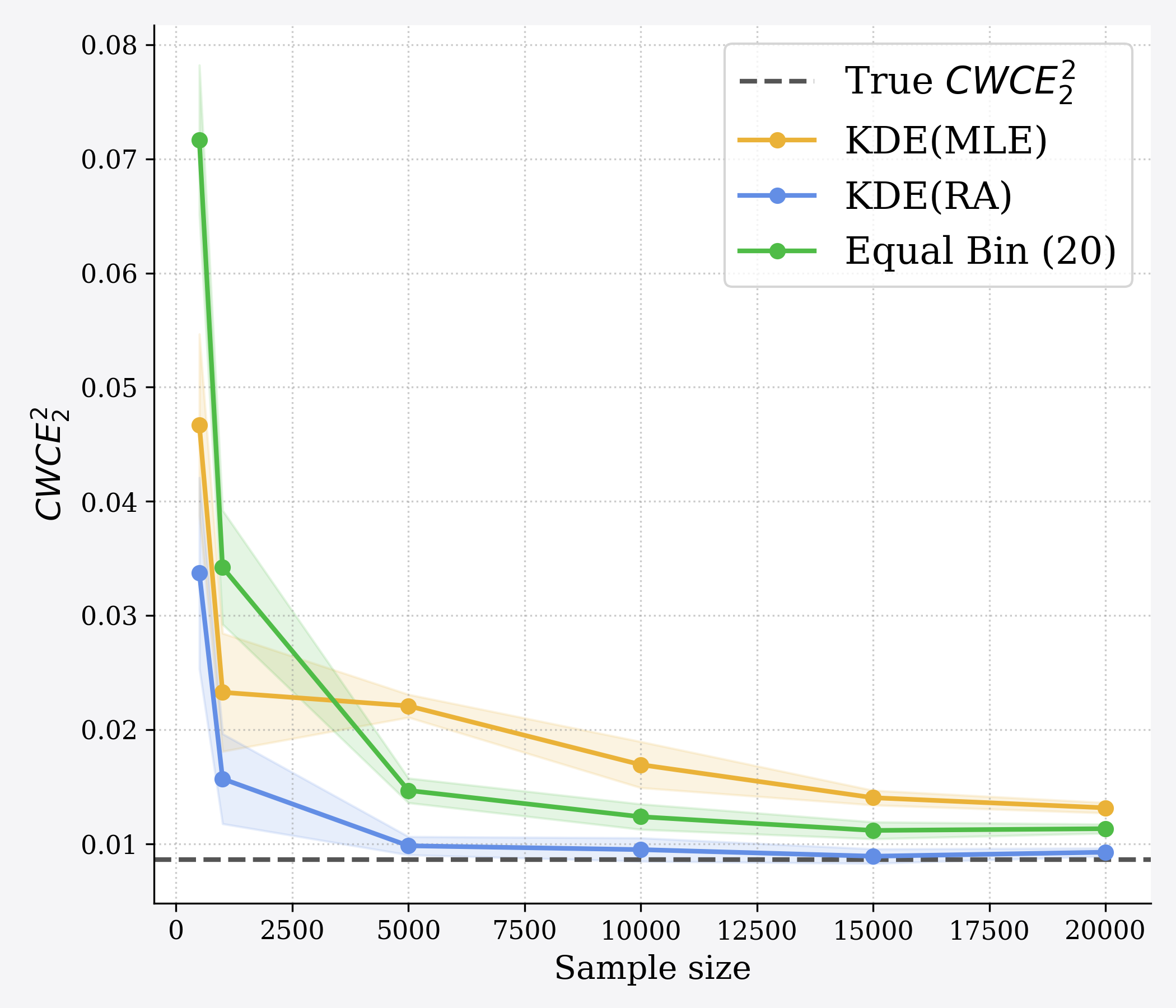}
    \subcaption{\textbf{$\text{CWCE}_2^2$}}
    \label{fig:compare_ce}
\end{minipage}%
\hspace{0.01\linewidth}  
\begin{minipage}[t]{1.0\linewidth} 
    \centering
    \includegraphics[width=\linewidth]{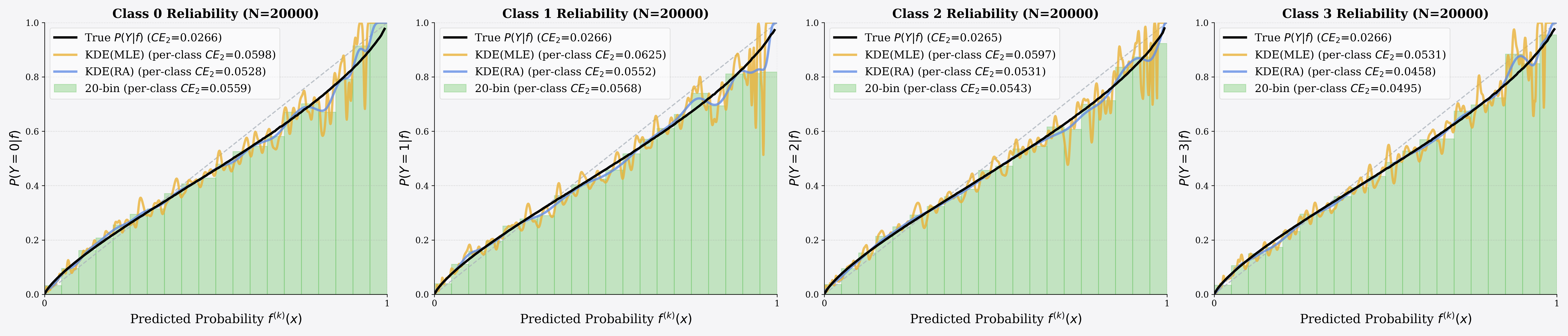}
    \subcaption{ \textbf{Reliability}}
    \label{fig:compare_rc}
\end{minipage}%
\caption{\footnotesize{Bandwidth selection behavior on synthetic data. \ref{fig:mle_bw} and \ref{fig:ra_bw} illustrates the results of $\hat{R}_i$ using the bandwidth $h$ selected by MLE or RA objective. While MLE is designed for density recovery, it tends to favor pathologically smaller bandwidths compared with RA objective. This preference leads to the behavior shown in the reliability plots (\ref{fig:mle_bw}, \ref{fig:ra_bw} and \ref{fig:compare_rc}), where the MLE-selected bandwidth causes the estimated bias of $\hat{R}_i$ to manifest as sharp spikes—reflecting overfitting—rather than the robust, smooth distribution produced by our Risk-based approach. Consequently, as demonstrated in \ref{fig:compare_ce}, this poorly tuned bandwidth results in suboptimal calibration error estimation.}}
\label{fig:mle}
\end{figure*}

To reveal the internal structural imbalance of risk-based alignment, we analyze the bias of the reconstructed risk at each individual sample $x_i$. For calibration error, we decompose its expectation of by centering $\hat{R}_i$ around the oracle value $R_i$:
\begin{align}
 \mathbb{E} [\widehat{\text{ce}}_i] = \mathbb{E} [ ((\hat{R}_i - R_i) + (R_i - f(x_i)))^2 ] 
\end{align}
Using the identity $\mathbb{E}[(\hat{R}_i - R_i)^2] = \text{Var}(\hat{R}_i) + (\mathbb{E}[\hat{R}_i] - R_i)^2$, the local calibration estimation bias relative to the oracle value $(R_i - f(x_i))^2$ is:
\begin{align}
\text{Bias}(\widehat{\text{ce}}_i)  = &  2(R_i - f(x_i))(\mathbb{E}[\hat{R}_i] - R_i) + \text{Var}(\hat{R}_i) \\
&+ (\mathbb{E}[\hat{R}_i] - R_i)^2.  \nonumber
\end{align}
Since the bias of the KDE estimator $\hat{R}_i$ is known to be $\mathcal{O}(h^2)$ \citep{dehnad1987density, wand1994kernel}, the squared bias term $(\mathbb{E}[\hat{R}_i] - R_i)^2$ is of order $\mathcal{O}(h^4)$. Neglecting this higher-order term, we obtain:
\begin{equation} \label{eq:local_ce_bias}
\scalebox{0.96}{$\displaystyle \text{Bias}(\widehat{\text{ce}}_i) \approx 2(R_i - f(x_i))(\mathbb{E}[\hat{R}_i] - R_i) + \text{Var}(\hat{R}_i).$}
\end{equation}
Similarly, for the refinement component $\widehat{\text{ref}}_i = \hat{R}_i(1 - \hat{R}_i)$, we analyze its expectation by utilizing the identity $\mathbb{E}[\hat{R}_i^2] = \text{Var}(\hat{R}_i) + (\mathbb{E}[\hat{R}_i])^2$:
\begin{align}
\mathbb{E} [\widehat{\text{ref}}_i] = \mathbb{E}[\hat{R}_i] - \mathbb{E}[\hat{R}_i^2] = \mathbb{E}[\hat{R}_i] - \left( \text{Var}(\hat{R}_i) + (\mathbb{E}[\hat{R}_i])^2 \right).  \nonumber
\end{align}
Subtracting the oracle refinement value $R_i(1 - R_i)$, the bias is given by:
\begin{align}
\text{Bias}(\widehat{\text{ref}}_i) = (\mathbb{E}[\hat{R}_i] - R_i) - \left( (\mathbb{E}[\hat{R}_i])^2 - R_i^2 \right) - \text{Var}(\hat{R}_i). \nonumber
\end{align}
Applying the identity $(\mathbb{E}[\hat{R}_i])^2 - R_i^2 = (\mathbb{E}[\hat{R}_i] - R_i)(\mathbb{E}[\hat{R}_i] + R_i)$ and substituting $\mathbb{E}[\hat{R}_i] + R_i = 2R_i + (\mathbb{E}[\hat{R}_i] - R_i)$, we expand the term:
\begin{align}
(\mathbb{E}[\hat{R}_i])^2 - R_i^2 &= (\mathbb{E}[\hat{R}_i] - R_i)(2R_i + \mathbb{E}[\hat{R}_i] - R_i) \nonumber \\
&= 2R_i(\mathbb{E}[\hat{R}_i] - R_i) + (\mathbb{E}[\hat{R}_i] - R_i)^2. \nonumber
\end{align}
Substituting this back into the refinement bias yields:
\begin{align}
\text{Bias}(\widehat{\text{ref}}_i) = &  (1 - 2R_i)(\mathbb{E}[\hat{R}_i] - R_i) - \text{Var}(\hat{R}_i) \\
& - (\mathbb{E}[\hat{R}_i] - R_i)^2. \nonumber 
\end{align}
Again, neglecting the $\mathcal{O}(h^4)$ term, we obtain the refinement bias:
\begin{align} \label{eq:local_ref_bias}
\text{Bias}(\widehat{\text{ref}}_i) \approx (1 - 2R_i)(\mathbb{E}[\hat{R}_i] - R_i) - \text{Var}(\hat{R}_i).
\end{align}
Equations~\eqref{eq:local_ce_bias} and~\eqref{eq:local_ref_bias} highlight a critical trade-off: individual estimators are contaminated by the variance of the estimator of $R$. In the calibration term, this variance acts as a spurious positive bias, whereas in the refinement term, it manifests as a negative bias. Summing these components leads to the exact cancellation of the second-order terms, yielding the risk bias:
\begin{align} \label{eq:local_risk_bias}
\text{Bias}(\hat{r}_{\text{kde},i}) \approx w_i (\mathbb{E}[\hat{R}_i] - R_i),
\end{align}
where $w_i = 1 - 2f(x_i)$ for the binary $L_2$ case. 
Thus the bias of the aggregate reconstructed risk in Eq. (15) is given by:
\begin{align} \label{eq:aggregate_risk_bias}
\text{Bias}\left(\sum_{i=1}^n \hat{r}_{\text{kde},i}\right) \approx  \sum_{i=1}^n w_i (\mathbb{E}[\hat{R}_i] - R_i).
\end{align}
This internal cancellation of variance terms is central to the proposed risk alignment objective. We demonstrate that this property generalizes across various classification settings and proper scoring rules in Appendix~\ref{appendix:extensions}. The functional forms of the resulting weights $w_i$ are summarized in Table~\ref{tab:bias_weights}, with complete derivations provided in the Appendix.

\begin{table}[htbp]
\centering
\caption{Weights $w_i$ for various proper scoring rules.}
\label{tab:bias_weights}
\addtolength{\tabcolsep}{0.8pt} 
\begin{tabular}{lll}
\toprule
\textbf{Scoring Rule}        & \textbf{Setting} & \textbf{Weight ($w_i$)} \\ 
\midrule
\textbf{Brier Score ($L_2$)} & Binary           & $1 - 2f(x_i)$                   \\
                             & Multi-class       & $-2f^{(k)}(x_i)$               \\
                             & Class-wise       & $1 - 2f^{(k)}(x_i)$            \\ 
\midrule
\textbf{Log Loss (KL)}       & Binary           & $\log \frac{1-f(x_i)}{f(x_i)}$  \\ \addlinespace[4pt]
                             & Multi-class      & $-\log f^{(k)}(x_i)$            \\ \addlinespace[4pt]
                             & Class-wise       & $\log \frac{1-f^{(k)}(x_i)}{f^{(k)}(x_i)}$ \\ 
\bottomrule
\end{tabular}
\end{table}

\paragraph{Variance-Bias Trap.} 
 The RA objective exploits the structural stability of the total risk to identify an optimal bandwidth. While MLE is a standard criterion for density estimation, it frequently fails in calibration because its objective is maximized when the estimated density peaks at observed sample locations. As $n$ increases, this causes the kernel to contract around individual data points, where the KDE estimator $\hat{R}_i$ degenerates into a nearest-neighbor label sampler. Although this fits local density, it maximizes the estimation variance $\text{Var}(\hat{R}_i)$. As revealed by Eq.~\eqref{eq:local_ce_bias}, the individual calibration estimator $\widehat{\text{ce}}_i$ is biased upward by this variance, which explains the failure of MLE-selected bandwidths. In contrast, Eq.~\eqref{eq:local_risk_bias} demonstrates that the total risk $\hat{r}_{\text{kde},i}$ remains robust because $\text{Var}(\hat{R}_i)$ is neutralized by the refinement component $\widehat{\text{ref}}_i$ through internal cancellation. This mechanism explains the stability of classical binning: by averaging local labels, it acts as a primitive low-pass filter that mitigates estimation variance, albeit while introducing discretization bias. By utilizing the RA objective, we achieve a similar suppression of local stochastic noise while preserving the smoothness of the estimator. As shown in Eq.~\eqref{eq:aggregate_risk_bias}, the aggregate risk bias remains coupled with the estimation bias of $R_i$ only through the weight $w_i$. By decoupling the bias from variance terms, the RA framework circumvents the bias-variance tradeoff and recovers the underlying calibration mapping, as illustrated in Fig.~\ref{fig:ra_bw}.

\section{Experiments}

\subsection{Synthetic data} \label{sec:synthetic}
Following \citet{popordanoska2022consistent}, we evaluate different calibration error estimators on synthetic data where the oracle $R^\star_i= \mathbb{E}[y\mid f(x_i)]$ is known by design, allowing for exact calibration error quantification.

\paragraph{Data generation.} We simulate $N=20,000$ i.i.d. samples for $K$ classes. For each instance, a latent vector $u_i \in \Delta^{K-1}$ is sampled uniformly via the Kraemer algorithm \citep{smith2004sampling}. We define the oracle distribution as $R_i^\star = \sigma(\log u_i / t_1)$, from which labels are drawn $y_i \sim \mathrm{Categorical}(R_i^\star)$ to ensure $\mathbb{E}[y \mid f(x)] = R^\star$. Miscalibration is then induced by setting the model output to $f(x_i) = \mathrm{softmax}(\log R_i^\star / t_2)$. We set $t_1 = 1.0$ and $t_2 = 0.8$ throughout.

\paragraph{Ground-truth calibration error.}Since $R^\star$ and $f(x)$ are known, the ground-truth canonical calibration error (Eq.~\eqref{eq:riskdecompose}) can be computed over the dataset as $ \mathrm{CE}^{\star}_{F}(f) =\frac{1}{N}\sum_{i=1}^{N} D_F\bigl(R^\star_i, f(x_i)\bigr). $ Analogously, the ground-truth classwise calibration error $\mathrm{CWCE}_F^{\star}(f)$ is evaluated by applying the same principle across all classes.

\paragraph{Baselines.} We evaluate KDE-based calibration error estimators through different bandwidth selection strategies, such as MLE and our RA method. We set $p=2$ for RA objective in Eq.~\eqref{eq:risk_opt_obj}. Both employ a grid search across the same domain $\mathcal{H}$ to ensure a fair comparison. Specifically, the bandwidth $h$ is optimized per class in classwise settings, whereas in canonical settings, a single global bandwidth is optimized. For binning-based baselines, we adopt the joint simplex binned estimator for canonical calibration~\citep{popordanoska2022consistent}, which partitions the $(K-1)$-dimensional probability simplex into a fixed-width grid to approximate the oracle $R_i^\star$. Consistent with \citet{popordanoska2022consistent}, we employ leave-one-out (LOO) unbiased estimation within each partition. For the $L_2$ setting, we additionally compare against the Kernel Ridge Regression-based (KRR) estimator~\citep{gruber2025optimizing}. For classwise calibration, we adopt the adaptive binning scheme~\citep{ding2020revisiting}, which adjusts bin edges based on class-specific confidence distributions.

\begin{figure*}[htbp] 
\centering 
\includegraphics[width=1.0\linewidth]{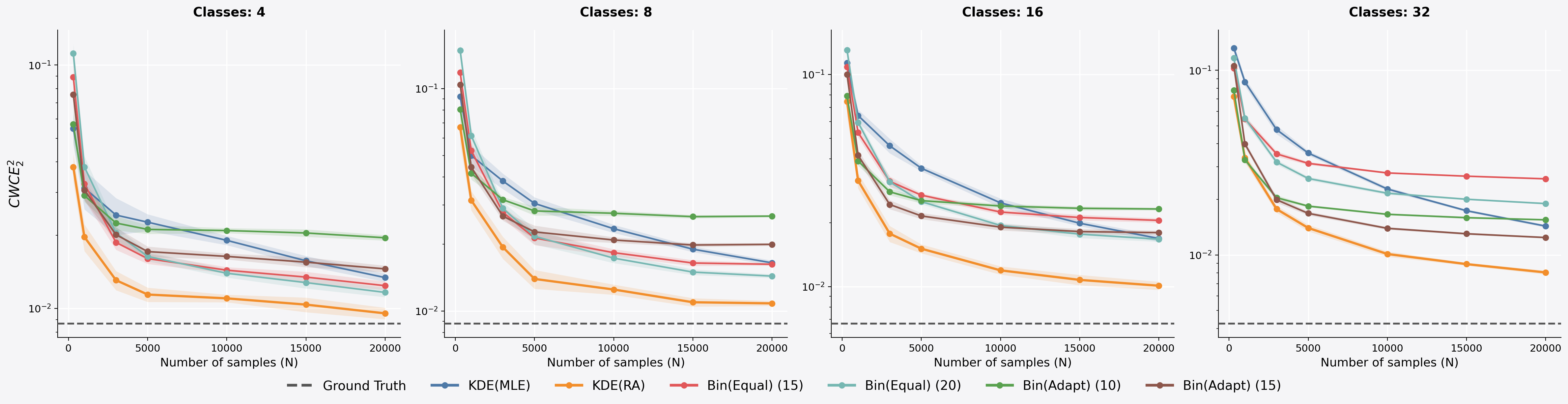}
\caption{(\textbf{Synthetic data}) Finite sample estimator of CWCE$_2^2$: mean and 95\% confidence interval computation across all subsamples of synthetic data.}
\label{fig:syn_classwise_l2}
\end{figure*}

\begin{figure*}[htbp] 
\centering 
\includegraphics[width=1.0\linewidth]{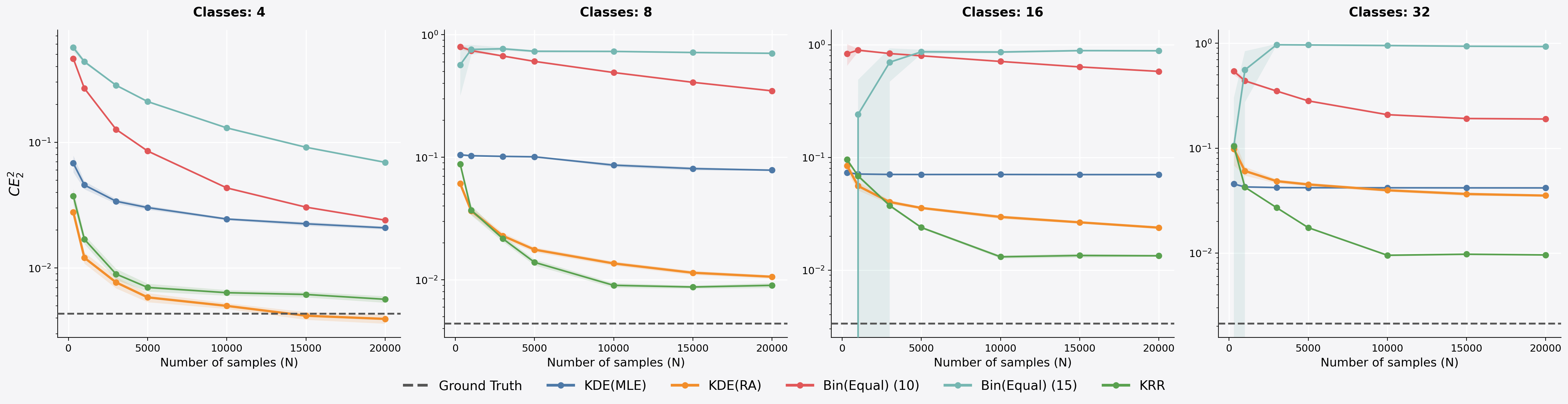}
\caption{(\textbf{Synthetic data}) Finite sample estimator of CE$_2^2$: mean and 95\% confidence interval computation across all subsamples of synthetic data.}
\label{fig:syn_canonical_l2}
\end{figure*}

\paragraph{Results.} Fig. \ref{fig:syn_classwise_l2} (Appendix Fig. \ref{fig:syn_classwise_kl}) and Fig. \ref{fig:syn_canonical_l2} (Appendix Fig. \ref{fig:syn_canonical_kl}) illustrate the performance of the classwise estimator $\widehat{\mathrm{CWCE}}_F(f)$ and canonical estimator $\widehat{\mathrm{CE}}_F(f)$, respectively, as a function of $n$. These results are commputed over 10 subsamples for scenarios involving 4, 8, 16, and 32 classes. As shown in Fig.~\ref{fig:syn_classwise_l2} and \ref{fig:syn_classwise_kl}, the KDE estimator utilizing the RA objective achieves the best performance compared to all existing methods in the classwise settings. Furthermore, the reliability plots in Fig.~\ref{fig:compare_rc} demonstrate that our proposed method aligns more closely with the ground truth $R^\star$ than either the MLE-based or binning approaches. As illustrated in Fig. \ref{fig:syn_canonical_l2} and \ref{fig:syn_canonical_kl}, binning methods demonstrate poor performance in canonical cases, significantly underperforming KDE methods. Despite outperforming MLE-based KDE, KDE-based estimators are sensitive to the increasing dimensionality of the probability simplex in the canonical setting. As the number of classes $K$ grows, the resulting data sparsity on the high-dimensional simplex makes it increasingly difficult for the Dirichlet kernel to accurately fit the underlying distribution. Thus KDE methods become less competitive than KKR in that setting.

\begin{table*}[htbp]
\caption{Summary of MAE of $\text{CWCE}_2^2 \times 10^{2}$ ($\downarrow$) with standard deviations. Best values are in \textcolor{best}{red}, and second-best in \textcolor{second}{teal}. Results are reported at $N=20,000$ for CIFAR-10/100, Amazon and ImageNet.}
\centering
\renewcommand{\arraystretch}{1.1}
\setlength{\tabcolsep}{9pt}
\resizebox{\textwidth}{!}{
\begin{tabular}{@{}llcccccc@{}}
\toprule
\multirow{2}{*}{\textbf{Dataset}} & \multirow{2}{*}{\textbf{Model}} & \multicolumn{2}{c}{\textbf{Equal Binning}} & \multicolumn{2}{c}{\textbf{Adaptive Binning}} & \multicolumn{2}{c}{\textbf{KDE Methods}} \\
\cmidrule(lr){3-4} \cmidrule(lr){5-6} \cmidrule(lr){7-8}
& & \textbf{15} & \textbf{20} & \textbf{10} & \textbf{15} & \textbf{MLE} & \textbf{RA} \\
\midrule
\multirow{4}{*}{\textbf{CIFAR10}} 
& PreResNet-20 & \textcolor{second}{0.19$_{\pm 0.03}$} & 0.28$_{\pm 0.05}$ & 1.98$_{\pm 0.58}$ & 30.61$_{\pm 0.09}$ & 1.15$_{\pm 0.08}$ & \textcolor{best}{0.07$_{\pm 0.05}$} \\
& PreResNet-56 & \textcolor{second}{0.19$_{\pm 0.05}$} & 0.30$_{\pm 0.05}$ & 1.98$_{\pm 0.70}$ & 31.44$_{\pm 0.08}$ & 1.10$_{\pm 0.09}$ & \textcolor{best}{0.08$_{\pm 0.04}$} \\
& VGG-16 (BN)  & \textcolor{best}{0.06$_{\pm 0.03}$} & 0.16$_{\pm 0.04}$ & 1.99$_{\pm 0.66}$ & 31.92$_{\pm 0.06}$ & 1.15$_{\pm 0.07}$ & \textcolor{second}{0.10$_{\pm 0.08}$} \\
& WRN-28-10    & \textcolor{second}{0.20$_{\pm 0.02}$} & 0.25$_{\pm 0.02}$ & 1.80$_{\pm 0.37}$ & 30.81$_{\pm 0.08}$ & 0.88$_{\pm 0.03}$ & \textcolor{best}{0.08$_{\pm 0.02}$} \\
\midrule
\multirow{4}{*}{\textbf{CIFAR100}} 
& PreResNet-20 & \textcolor{second}{1.55$_{\pm 0.11}$} & 2.15$_{\pm 0.09}$ & 129.58$_{\pm 0.26}$ & 119.85$_{\pm 0.26}$ & 9.03$_{\pm 0.16}$ & \textcolor{best}{0.85$_{\pm 0.08}$} \\
& PreResNet-56 & \textcolor{second}{1.57$_{\pm 0.08}$} & 2.03$_{\pm 0.09}$ & 167.13$_{\pm 0.17}$ & 157.14$_{\pm 0.17}$ & 5.78$_{\pm 0.09}$ & \textcolor{best}{0.85$_{\pm 0.08}$} \\
& VGG-16 (BN)  & \textcolor{best}{0.36$_{\pm 0.06}$} & 0.83$_{\pm 0.06}$ & 172.27$_{\pm 0.11}$ & 162.28$_{\pm 0.12}$ & 4.76$_{\pm 0.15}$ & \textcolor{second}{0.82$_{\pm 0.14}$} \\
& WRN-28-10    & \textcolor{second}{1.41$_{\pm 0.08}$} & 1.87$_{\pm 0.05}$ & 171.58$_{\pm 0.14}$ & 161.60$_{\pm 0.14}$ & 4.79$_{\pm 0.17}$ & \textcolor{best}{0.61$_{\pm 0.10}$} \\
\midrule
\multirow{4}{*}{\textbf{AMAZON}} 
& Bert            & 2.07$_{\pm 0.45}$ & 1.83$_{\pm 0.42}$ & 4.83$_{\pm 0.47}$ & 3.38$_{\pm 0.45}$ & \textcolor{second}{0.85$_{\pm 0.50}$} & \textcolor{best}{0.38$_{\pm 0.29}$} \\
& Distill Bert    & 1.18$_{\pm 0.31}$ & 1.04$_{\pm 0.33}$ & 3.93$_{\pm 0.29}$ & 2.57$_{\pm 0.33}$ & \textcolor{second}{0.99$_{\pm 0.32}$} & \textcolor{best}{0.23$_{\pm 0.25}$} \\
& Roberta         & 0.31$_{\pm 0.25}$ & \textcolor{best}{0.23$_{\pm 0.20}$} & 3.80$_{\pm 0.22}$ & 2.27$_{\pm 0.26}$ & 0.95$_{\pm 0.30}$ & \textcolor{second}{0.27$_{\pm 0.15}$} \\
& Distill Roberta & 1.65$_{\pm 0.26}$ & 1.43$_{\pm 0.26}$ & 4.23$_{\pm 0.34}$ & 2.96$_{\pm 0.30}$ & \textcolor{second}{1.10$_{\pm 0.30}$} & \textcolor{best}{0.23$_{\pm 0.15}$} \\
\midrule
\multirow{4}{*}{\textbf{IMAGENET}} 
& ViT-Small  & \textcolor{second}{11.24$_{\pm 0.26}$} & 14.16$_{\pm 0.29}$ & 125.55$_{\pm 0.20}$ & 124.56$_{\pm 0.20}$ & 34.03$_{\pm 0.21}$ & \textcolor{best}{5.45$_{\pm 0.35}$} \\
& ViT-Base   & \textcolor{second}{9.86$_{\pm 0.19}$}  & 12.32$_{\pm 0.22}$ & 121.47$_{\pm 0.22}$ & 120.62$_{\pm 0.22}$ & 28.76$_{\pm 0.34}$ & \textcolor{best}{4.27$_{\pm 0.24}$} \\
& Swin-Small & \textcolor{second}{9.30$_{\pm 0.11}$}  & 11.68$_{\pm 0.16}$ & 117.68$_{\pm 0.19}$ & 116.85$_{\pm 0.19}$ & 27.11$_{\pm 0.36}$ & \textcolor{best}{4.29$_{\pm 0.24}$} \\
& DeiT-Base  & \textcolor{second}{9.08$_{\pm 0.15}$}  & 11.48$_{\pm 0.19}$ & 114.44$_{\pm 0.21}$ & 113.68$_{\pm 0.21}$ & 27.22$_{\pm 0.23}$ & \textcolor{best}{3.77$_{\pm 0.06}$} \\
\bottomrule
\end{tabular}}
\label{tab:comprehensive_mae_colored}
\end{table*}

\subsection{Real-world data}
\paragraph{Datasets.} Evaluation is conducted on image classification benchmarks, including CIFAR-10/100~\citep{krizhevsky2009learning} and ImageNet~\citep{deng2009imagenet}, alongside the Amazon Reviews text dataset~\citep{ni2019justifying}. The Amazon dataset comprises review text paired with five-star ordinal ratings as labels. For CIFAR-10/100, we aggregate the training and test sets to serve as the population, whereas for ImageNet, we utilize the validation set. 

\paragraph{Models.} For experiments on CIFAR10/100, we report the results on the VGG16~\citep{simonyan2014very} model with batch norm layers, WideResNet28x10~\citep{zagoruyko2016wide}, and ResNet~\citep{he2016deep} models with different depths $(20, 56)$. We evaluate the calibration error estimators using models that are pre-trained on the CIFAR10/100 dataset. For experiments on Amazon dataset, we use pre-trained transformer-based models – BERT~\citep{devlin2018bert}, RoBERTa~\citep{liu2021robustly}, Distill-Bert (D-BERT), and Distill-Roberta (D-RoBERTa)~\citep{sanh2019distilbert}. For experiments on the ImageNet dataset, we use the pre-trained models from \textit{timm}~\citep{rw2019timm} package, including two Vision Transformer (ViT) variants~\citep{dosovitskiy2020image}, ViT-Small and ViT-Base, along with a Swin Transformer-Small~\citep{liu2021swin} and a Data-efficient Image Transformer (DeiT-Base)~\citep{touvron2021training}. 

\update{\paragraph{Proxy ground-truth calibration error.} For real datasets, the ground truth cannot be directly derived and must be inferred. We therefore adopted the best-performing estimators identified from our synthetic experiments—KRR for canonical $L_2$ error and RA for all other metrics. Crucially, we computed these proxies over the largest available data pools (e.g., the combined training and test sets for CIFAR-10/100). The theoretical validity of our KDE method is supported by \cite{ouimet2022asymptotic}, who establishes that Dirichlet KDE achieves optimal convergence under specific bandwidth constraints. Satisfying these constraints depends heavily on dimensionality. At our population scale, 1D classwise calibration strictly meets these conditions, yielding a mathematically robust proxy. Conversely, because canonical calibration in high-dimensional spaces (e.g., 1,000 ImageNet classes) is statistically prohibitive due to the curse of dimensionality, we exclude it from our ImageNet evaluation.}

\paragraph{Experimental setup.}To assess estimator performance across varying sample sizes, we report the mean and 95\% confidence intervals calculated over 10 independent subsets. Additionally, the standard deviations are derived from those subsets by measuring the absolute error of each estimate relative to the ground truth. Additional results can be found in Appendix~\ref{appendix:addtionalresults}. 

\paragraph{Results.} Consistent with our synthetic findings, the RA method demonstrate a clear superiority over the MLE-based approach, particularly in estimating class-wise calibration error as shown in Table~\ref{tab:comprehensive_mae_colored}. In contrast, among binning techniques, the adaptive binning method underperforms the equal-width binning method in class-wise settings. This underperformance is likely due to the overconfidence of pre-trained networks; the resulting concentration of probabilities near $1$ forces adaptive binning to create extremely narrow bins at the extremes and overly coarse bins in the middle. Consequently, this non-uniform resolution fails to capture calibration gaps in sparse regions, whereas equal-width binning maintains a stable and representative resolution across the entire unit interval.

\section{Discussion and Conclusion}
In this paper, we revisited the fundamental challenge of bandwidth selection for KDE-based calibration error estimators. While KDE offers a differentiable and smooth alternative to traditional binning, its reliability is notoriously sensitive to the kernel bandwidth. The MLE objective often collapses toward pathologically small bandwidths, causing the estimator to overfit to local noise. 

To address this, we introduce Risk Alignment (RA) objective, a novel optimization framework grounded in the Bregman decomposition of proper scoring rules. Our theoretical analysis demonstrates that while the biases of individual calibration and refinement estimators are coupled with the estimation variance of the conditional probability $R$, their sum—the total risk—enables a systematic variance cancellation. This property ensures that the RA objective is decoupled from high-variance artifacts and is primarily driven by the bias of the conditional probability, scaled by a confidence-dependent weight. By overcoming the inherent variance-bias tradeoff in calibration estimation, RA facilitates a robust recovery of the conditional probability manifold. Extensive experiments show that RA consistently outperforms MLE in calibration estimation, particularly in class-wise settings.

Despite its advantages, our study identifies several remaining challenges. Experiments show that KDE methods, including RA, can degrade in extremely high-dimensional settings, such as canonical calibration with many classes, primarily due to data sparsity on the simplex. In these regimes, the Dirichlet kernel may be less effective than the RBF kernel, suggesting that while RA optimizes bandwidth, the choice of kernel remains critical. Since RA is grounded in Bregman decomposition, its objective can be readily adapted to alternative kernels like RBF to mitigate high-dimensional issues. Furthermore, integrating RA-optimized KDE as a differentiable loss term is a promising direction. By providing stable and less biased gradients, RA could facilitate more effective calibration-aware training.

\begin{highlight}
\section*{Acknowledgments}
This research received funding from the Flemish Government (AI Research Program) and the Research Foundation
- Flanders (FWO) through project number G0G2921N. HZ is supported by the China Scholarship Council. We acknowledge EuroHPC JU for awarding the project ID EHPC-DEV-2025D06-055 and EHPC-DEV-2025D12-054 access to the EuroHPC supercomputer LEONARDO, hosted by CINECA (Italy) and the LEONARDO consortium.
\end{highlight}

\bibliography{ref}

\newpage

\onecolumn

\title{Bandwidth Selection in Kernel Density Estimation for Model Calibration (Supplementary Material)}
\maketitle

\appendix

\section{Appendix: Extension to Multi-class and KL Divergence} \label{appendix:extensions}

While the main text focuses on the binary $L_2$ case for clarity, the risk-consistency principle is fundamentally rooted in the properties of Bregman divergences. In this section, we demonstrate that the internal cancellation of estimation variance generalizes to multi-class settings and alternative proper scoring rules, such as the KL divergence (Cross-Entropy).

\subsection{Multi-class Canonical $L_2$ Risk (Brier Score)}
We consider a model output $f(x)$ on the $K$-dimensional unit simplex $\Delta^{K-1}$. Let $f^{(k)}(x_i) \in [0,1]$ be the predicted probability for class $k \in \{1, \dots, K\}$ and $y^{(k)} \in \{0, 1\}$ be the ground truth label. 

\paragraph{Canonical $L_2$ Bregman Decomposition:}
Following the quadratic properties of the $L_2$ scoring rule, the multi-class Brier score admits an orthogonal decomposition:
\begin{align}
\mathcal{R}_{L_2}(f) = \underbrace{\mathbb{E} \left[ \sum_{k=1}^K (R^{(k)} - f^{(k)}(x))^2 \right]}_{\text{CE}_{2}^2(f)} + \underbrace{\mathbb{E} \left[ \sum_{k=1}^K R^{(k)}(1 - R^{(k)}) \right]}_{\text{REF}_{2}(f)},
\end{align}
where $R^{(k)}$ denotes the true conditional probability $P(Y=k|f(x))$. We reconstruct the pointwise risk as $\hat{r}_{\text{kde},i} = \sum_{k=1}^K (\widehat{\text{ce}}_i^{(k)} + \widehat{\text{ref}}_i^{(k)})$, where $\widehat{\text{ce}}_i^{(k)} = (\hat{R}_i^{(k)} - f^{(k)}(x_i))^2$ and $\widehat{\text{ref}}_i^{(k)} = \hat{R}_i^{(k)}(1 - \hat{R}_i^{(k)})$.

\paragraph{Multi-class Bias Derivation:}
To analyze the stability of this estimator, we consider the bias for a single sample $i$ and class $k$. Centering the calibration term around the oracle value $R_i^{(k)}$:
\begin{align}
\mathbb{E} \left[ \widehat{\text{ce}}_i^{(k)} \right] = \mathbb{E} \left[ (\hat{R}_i^{(k)} - R_i^{(k)})^2 + 2(\hat{R}_i^{(k)} - R_i^{(k)})(R_i^{(k)} - f^{(k)}(x_i)) + (R_i^{(k)} - f^{(k)}(x_i))^2 \right].
\end{align}
Using the identity $\mathbb{E}[(\hat{R}_i^{(k)} - R_i^{(k)})^2] = \text{Var}(\hat{R}_i^{(k)}) + (\mathbb{E}[\hat{R}_i^{(k)}] - R_i^{(k)})^2$, the bias is:
\begin{align}
\text{Bias}\bigl(\widehat{\text{ce}}_i^{(k)}\bigr) = 2(R_i^{(k)} - f^{(k)}(x_i))(\mathbb{E}[\hat{R}_i^{(k)}] - R_i^{(k)}) + \text{Var}(\hat{R}_i^{(k)}) + (\mathbb{E}[\hat{R}_i^{(k)}] - R_i^{(k)})^2.
\end{align}
Neglecting the $\mathcal{O}(h^4)$ squared bias term $(\mathbb{E}[\hat{R}_i^{(k)}] - R_i^{(k)})^2$ \citep{dehnad1987density}, the bias of the calibration error estimator is:
\begin{align} \label{eq:appendix_l2_ce_bias_derivation}
\text{Bias}\bigl(\widehat{\text{ce}}_i^{(k)}\bigr) \approx 2(R_i^{(k)} - f^{(k)}(x_i))(\mathbb{E}[\hat{R}_i^{(k)}] - R_i^{(k)}) + \text{Var}(\hat{R}_i^{(k)}).
\end{align}
Similarly, for the refinement component, expanding $\mathbb{E}[(\hat{R}_i^{(k)})^2] = \text{Var}(\hat{R}_i^{(k)}) + (\mathbb{E}[\hat{R}_i^{(k)}])^2$ and applying the difference of squares $(\mathbb{E}[\hat{R}_i^{(k)}])^2 - (R_i^{(k)})^2 = (\mathbb{E}[\hat{R}_i^{(k)}] - R_i^{(k)})(\mathbb{E}[\hat{R}_i^{(k)}] + R_i^{(k)})$, we have:
\begin{align}
\text{Bias}\bigl(\widehat{\text{ref}}_i^{(k)}\bigr) &= (\mathbb{E}[\hat{R}_i^{(k)}] - R_i^{(k)}) - \left( (\mathbb{E}[\hat{R}_i^{(k)}])^2 - (R_i^{(k)})^2 \right) - \text{Var}(\hat{R}_i^{(k)}) \nonumber \\
&= (1 - (\mathbb{E}[\hat{R}_i^{(k)}] + R_i^{(k)}))(\mathbb{E}[\hat{R}_i^{(k)}] - R_i^{(k)}) - \text{Var}(\hat{R}_i^{(k)}) \nonumber \\
&= (1 - (2R_i^{(k)} + \mathbb{E}[\hat{R}_i^{(k)}] - R_i^{(k)}))(\mathbb{E}[\hat{R}_i^{(k)}] - R_i^{(k)}) - \text{Var}(\hat{R}_i^{(k)}).
\end{align}
Neglecting the $\mathcal{O}(h^4)$ term, the refinement bias is:
\begin{align} \label{eq:appendix_l2_ref_bias_derivation}
\text{Bias}\bigl(\widehat{\text{ref}}_i^{(k)}\bigr) \approx (1 - 2R_i^{(k)})(\mathbb{E}[\hat{R}_i^{(k)}] - R_i^{(k)}) - \text{Var}(\hat{R}_i^{(k)}).
\end{align}
Summing these across all classes $k$ for each sample $i$, both the variance terms $\text{Var}(\hat{R}_i^{(k)})$ and the  $\mathcal{O}(h^4)$ cancel out exactly. The resulting pointwise risk bias is:
\begin{align}
\text{Bias}(\hat{r}_{\text{kde},i}) = \sum_{k=1}^K \left[ (1 - 2f^{(k)}(x_i)) \right] (\mathbb{E}[\hat{R}_i^{(k)}] - R_i^{(k)}).
\end{align}
By applying the simplex constraint $\sum_k R_i^{(k)} = 1$ and the model constraint $\sum_k f^{(k)}(x_i) = 1$, we expand the summation:
\begin{align}
\text{Bias}(\hat{r}_{\text{kde},i}) &\approx \left( \sum_{k=1}^K \mathbb{E}[\hat{R}_i^{(k)}] - \sum_{k=1}^K R_i^{(k)} \right) - 2 \sum_{k=1}^K f^{(k)}(x_i) (\mathbb{E}[\hat{R}_i^{(k)}] - R_i^{(k)}) \nonumber \\
&= \left( \sum_{k=1}^K \mathbb{E}[\hat{R}_i^{(k)}] - 1 \right) - 2 \sum_{k=1}^K f^{(k)}(x_i) (\mathbb{E}[\hat{R}_i^{(k)}] - R_i^{(k)}).
\end{align}
Under the assumption that the KDE estimator preserves the sum-to-one constraint, i.e., $\sum_k \mathbb{E}[\hat{R}_i^{(k)}] = 1$, the first term vanishes, and the pointwise risk bias simplifies to:
\begin{align} \label{eq:appendix_l2_total_bias_simplified}
 \text{Bias}(\hat{r}_{\text{kde},i}) \approx -2 \sum_{k=1}^K f^{(k)}(x_i) (\mathbb{E}[\hat{R}_i^{(k)}] - R_i^{(k)}).
\end{align}

\subsection{Binary KL Risk}
In the context of Kullback-Leibler (KL) divergence, we initially focus our analysis on the foundational case of binary classification for a predicted confidence $f(x) \in [0, 1]$ and ground-truth label $y \in \{0, 1\}$.

\paragraph{Binary KL Bregman Decomposition:}
The expected KL risk admits the following decomposition:
\begin{align} 
\mathcal{R}_{\text{KL}}(f) = \underbrace{\mathbb{E} [H(R)]}_{\text{REF}_{\text{KL}}(f)} + \underbrace{\mathbb{E} [D_{\text{KL}}(R, f(x))]}_{\text{CE}_{\text{KL}}(f)}, 
\end{align}
where $H(R) = -R \log R - (1-R) \log(1-R)$ is the refinement component. We reconstruct the pointwise risk as $\hat{r}_{\text{kde},i}(h) = \widehat{\text{ce}}_i(h) + \widehat{\text{ref}}_i(h)$, using the LOO-KDE estimates $\hat{R}_i$.

\paragraph{Binary Bias Derivation:}We analyze the bias via a second-order Taylor expansion around the true conditional probability $R_i$. For the calibration component $\widehat{\text{ce}}_i$, let $g(R) = R \log \frac{R}{f(x_i)} + (1-R) \log \frac{1-R}{1-f(x_i)}$. Its derivatives are:
\begin{align}g'(R) = \log \frac{R(1-f(x_i))}{f(x_i)(1-R)}, \quad g''(R) = \frac{1}{R(1-R)}.\end{align}The Taylor expansion $\mathbb{E}[g(\hat{R}_i)] \approx g(R_i) + g'(R_i)(\mathbb{E}[\hat{R}_i] - R_i) + \frac{1}{2}g''(R_i)\mathbb{E}[(\hat{R}_i - R_i)^2]$ yields the bias:
\begin{align} \label{eq:kl_ce_bias_binary_full}\text{Bias}\bigl(\widehat{\text{ce}}_i\bigr) \approx g'(R_i) (\mathbb{E}[\hat{R}_i] - R_i) + \frac{\text{Var}(\hat{R}_i) + (\mathbb{E}[\hat{R}_i] - R_i)^2}{2R_i(1-R_i)}.
\end{align}
Similarly, for the refinement component $\widehat{\text{ref}}_i = H(\hat{R}_i) = -\hat{R}_i \log \hat{R}_i - (1-\hat{R}_i) \log(1-\hat{R}_i)$, the derivatives are $H'(R) = \log \frac{1-R}{R}$ and $H''(R) = -\frac{1}{R(1-R)}$. The expansion yields:
\begin{align} \label{eq:kl_ref_bias_binary_full}\text{Bias}\bigl(\widehat{\text{ref}}_i\bigr) \approx H'(R_i) (\mathbb{E}[\hat{R}_i] - R_i) - \frac{\text{Var}(\hat{R}_i) + (\mathbb{E}[\hat{R}_i] - R_i)^2}{2R_i(1-R_i)}.
\end{align}
Since the bias of the KDE estimator $\hat{R}_i$ is known to be $\mathcal{O}(h^2)$ \citep{dehnad1987density, wand1994kernel}, the squared bias term $(\mathbb{E}[\hat{R}_i] - R_i)^2$ in both Equations~\eqref{eq:kl_ce_bias_binary_full} and~\eqref{eq:kl_ref_bias_binary_full} is of order $\mathcal{O}(h^4)$. Neglecting these higher-order terms, we obtain the calibration bias:
\begin{align} 
&\text{Bias}\bigl(\widehat{\text{ce}}_i\bigr) \approx \log \frac{R_i(1-f(x_i))}{f(x_i)(1-R_i)} (\mathbb{E}[\hat{R}_i] - R_i) + \frac{\text{Var}(\hat{R}_i)}{2R_i(1-R_i)}, \label{eq:kl_ce_bias_binary} \\
& \text{Bias}\bigl(\widehat{\text{ref}}_i\bigr) \approx \log \frac{1-R_i}{R_i} (\mathbb{E}[\hat{R}_i] - R_i) - \frac{\text{Var}(\hat{R}i)}{2R_i(1-R_i)}. \label{eq:kl_ref_bias_binary}
\end{align}
Standard KL-based calibration estimation is notoriously sensitive to the variance $\text{Var}(\hat{R}_i)$ near the boundaries. As shown in Eqs.~\eqref{eq:kl_ce_bias_binary} and~\eqref{eq:kl_ref_bias_binary}, the second-order bias coefficients diverge as $R_i \to 0$ or $1$, causing individual components to become numerically unstable. However, summing these equations leads to the exact cancellation of these boundary-sensitive terms. Combining the first-order terms, we have:
\begin{align}
\text{Bias}(\hat{r}_{\text{kde},i}) &\approx \left( \log \frac{R_i(1-f(x_i))}{f(x_i)(1-R_i)} + \log \frac{1-R_i}{R_i} \right) (\mathbb{E}[\hat{R}_i] - R_i) = \left(  \log \frac{1-f(x_i)}{f(x_i)} \right) (\mathbb{E}[\hat{R}_i] - R_i).\end{align}
The resulting pointwise risk bias is thus stabilized as a first-order functional.

\subsection{Multi-class Canonical KL Risk}
Extending the analysis to the multi-class setting, we consider a model output $f(x)$ on the $K$-dimensional unit simplex $\Delta^{K-1}$.

\paragraph{Multi-class KL Bregman Decomposition:}
The multi-class KL risk admits a decomposition into multi-class refinement and calibration components:
\begin{align}
\mathcal{R}_{\text{KL}}(f) = \underbrace{\mathbb{E} \left[ \sum_{k=1}^K -R^{(k)} \log R^{(k)} \right]}_{\text{REF}_{\text{KL}}(f)} + \underbrace{\mathbb{E} \left[ \sum_{k=1}^K R^{(k)} \log \frac{R^{(k)}}{f^{(k)}(x)} \right]}_{\text{CE}_{\text{KL}}(f)}.
\end{align}
We reconstruct the pointwise risk as $\hat{r}_{\text{kde},i} = \sum_k \widehat{\text{ce}}_i^{(k)} + \sum_k \widehat{\text{ref}}_i^{(k)}$.

\paragraph{Multi-class Bias Derivation:}
We perform a second-order multivariate Taylor expansion around $\mathbf{R}_i$. Let $\delta_i^{(k)} = \mathbb{E}[\hat{R}_i^{(k)}] - R_i^{(k)}$. For $g_{CE}(\mathbf{R}) = \sum_k R^{(k)} (\log R^{(k)} - \log f^{(k)})$, the partial derivative $\frac{\partial g_{CE}}{\partial R^{(k)}} = \log \frac{R^{(k)}}{f^{(k)}} + 1$, and Hessian entries $1/R^{(k)}$:
\begin{align} \label{eq:kl_ce_bias_multi_full}
\text{Bias}\bigl(\sum_k \widehat{\text{ce}}_i^{(k)}\bigr) \approx \sum_{k=1}^K \left[ \left( \log \frac{R_i^{(k)}}{f^{(k)}(x_i)} + 1 \right) \delta_i^{(k)} + \frac{\text{Var}(\hat{R}_i^{(k)}) + (\delta_i^{(k)})^2}{2R_i^{(k)}} \right].
\end{align}
For $g_{REF}(\mathbf{R}) = \sum_k -R^{(k)} \log R^{(k)}$, the derivatives are $-\log R^{(k)} - 1$ and Hessian entries $-1/R^{(k)}$:
\begin{align} \label{eq:kl_ref_bias_multi_full}
\text{Bias}\bigl(\sum_k \widehat{\text{ref}}_i^{(k)}\bigr) \approx \sum_{k=1}^K \left[ \left( -\log R_i^{(k)} - 1 \right) \delta_i^{(k)} - \frac{\text{Var}(\hat{R}_i^{(k)}) + (\delta_i^{(k)})^2}{2R_i^{(k)}} \right].
\end{align}
Summing Equations~\eqref{eq:kl_ce_bias_multi_full} and~\eqref{eq:kl_ref_bias_multi_full} and neglecting $\mathcal{O}(h^4)$ terms, the boundary-sensitive variance terms and $\log R_i^{(k)}$ terms cancel. Using the assumption that $\sum_k \delta_i^{(k)} = 0$ (simplex constraint preservation), the constant shifts also vanish:
\begin{align} \label{eq:canonical_classwise_kl}
\text{Bias}(\hat{r}_{\text{kde},i}) \approx \sum_{k=1}^K \left( \log R_i^{(k)} - \log f^{(k)}(x_i) + 1 - \log R_i^{(k)} - 1 \right) \delta_i^{(k)} = \sum_{k=1}^K \bigl( -\log f^{(k)}(x_i) \bigr) \delta_i^{(k)}.
\end{align}
\subsection{Class-wise One-versus-Rest Decomposition} \label{sec:classwiseOvR}

In addition to canonical calibration, it is often desirable to evaluate calibration on a per-class basis, especially in scenarios with class imbalance. Following \citet{popordanoska2024consistent}, the multi-class problem can be decomposed into $K$ independent binary tasks using a one-versus-rest (OvR) approach. We consider a model output $f(x)$ residing on the $(K-1)$-dimensional unit simplex $\Delta^{K-1}$. For each class $k \in \{1, \dots, K\}$, we denote the individual predicted probability as $f^{(k)}(x)$, which represents the model's confidence in class $k$ such that $\sum_{k=1}^K f^{(k)}(x) = 1$. The corresponding ground-truth label is represented by a one-hot vector $\mathbf{y}$, with binary indicators $y^{(k)} \in \{0, 1\}$ satisfying $\sum_{k=1}^K y^{(k)} = 1$.

The expected class-wise risk $\mathcal{R}_{\text{cw}}$ is the sum of these binary risks. For any proper scoring rule induced by $F$, the class-wise risk admits an individual Bregman decomposition for each class:
\begin{align}
\mathcal{R}_{\text{cw}}(f) = \sum_{k=1}^K \mathbb{E} [ \ell(f^{(k)}(x), y^{(k)}) ] = \sum_{k=1}^K \left( \text{CWCE}_F^{(k)}(f) + \text{REF}_F^{(k)}(f) \right).
\end{align}

\paragraph{Class-wise $L_2$ Calibration Error ($\text{CWCE}_2^2$):}
Under the $L_2$ loss, the total class-wise calibration error is defined as the sum of squared binary calibration errors: 
\begin{align} \label{eq:classwise_l2_ce}
\text{CWCE}_2^2(f)= \sum_{k=1}^K \mathbb{E} \left[ \bigl(R^{(k)} - f^{(k)}(x)\bigr)^2 \right].
\end{align}
Using the LOO-KDE estimates $\hat{R}_i^{(k)}$, the bias of the class-wise estimator follows the variance-cancellation mechanism derived in the binary $L_2$ case. For each class $k$, the positive variance term $\text{Var}(\hat{R}_i^{(k)})$ from the calibration component is neutralized by the corresponding negative term in the refinement component. By neglecting the $\mathcal{O}(h^4)$ squared bias terms for each class, the resulting total class-wise risk bias remains a stable first-order functional:
\begin{align} \label{eq:appendix_l2_cw_bias_final}
\text{Bias}\bigl(\widehat{\mathcal{R}}_{\text{cw}, L_2} (f) \bigr) &\approx \frac{1}{n} \sum_{i=1}^n \sum_{k=1}^K \left[ (2(R_i^{(k)} - f^{(k)}(x_i)) + (1 - 2R_i^{(k)})) (\mathbb{E}[\hat{R}_i^{(k)}] - R_i^{(k)}) \right] \nonumber \\
&= \frac{1}{n} \sum_{i=1}^n \sum_{k=1}^K (1 - 2f^{(k)}(x_i)) (\mathbb{E}[\hat{R}_i^{(k)}] - R_i^{(k)}).
\end{align}

\paragraph{Class-wise KL Calibration Error ($\text{CWCE}_{\text{KL}}$):}
Similarly, the class-wise calibration error induced by the KL divergence (Log loss) is the sum of binary KL divergences between the class-specific marginals: 
\begin{align}
\text{CWCE}_{\text{KL}}(f) =\sum_{k=1}^K \mathbb{E} \left[ D_{\text{KL}}\bigl(R^{(k)},f^{(k)}(x)\bigr) \right].
\end{align}
Applying the second-order Taylor expansion to each binary KL term, we observe that the boundary-sensitive variance terms $\text{Var}(\hat{R}_i^{(k)}) / (2R_i^{(k)}(1-R_i^{(k)}))$ cancel out exactly within each class $k$ when aggregated into the total risk. Neglecting higher-order terms, the stabilized bias for $\widehat{\mathcal{R}}_{\text{cw}, \text{KL}}$ is given by:
\begin{align} \label{eq:appendix_kl_cw_bias_final}
\text{Bias}\bigl(\widehat{\mathcal{R}}_{\text{cw}, \text{KL}} (f) \bigr) &\approx \frac{1}{n} \sum_{i=1}^n \sum_{k=1}^K \left[ \left( \log \frac{R_i^{(k)}(1-f^{(k)}(x_i))}{f^{(k)}(x_i)(1-R_i^{(k)})} + \log \frac{1-R_i^{(k)}}{R_i^{(k)}} \right) (\mathbb{E}[\hat{R}_i^{(k)}] - R_i^{(k)}) \right] \nonumber \\
&= \frac{1}{n} \sum_{i=1}^n \sum_{k=1}^K \left( \log \frac{1-f^{(k)}(x_i)}{f^{(k)}(x_i)} \right) \bigl( \mathbb{E}[\hat{R}_i^{(k)}] - R_i^{(k)} \bigr).
\end{align}

\section{Additional results} \label{appendix:addtionalresults}

\begin{figure*}[htbp] 
\centering 
\includegraphics[width=1.0\linewidth]{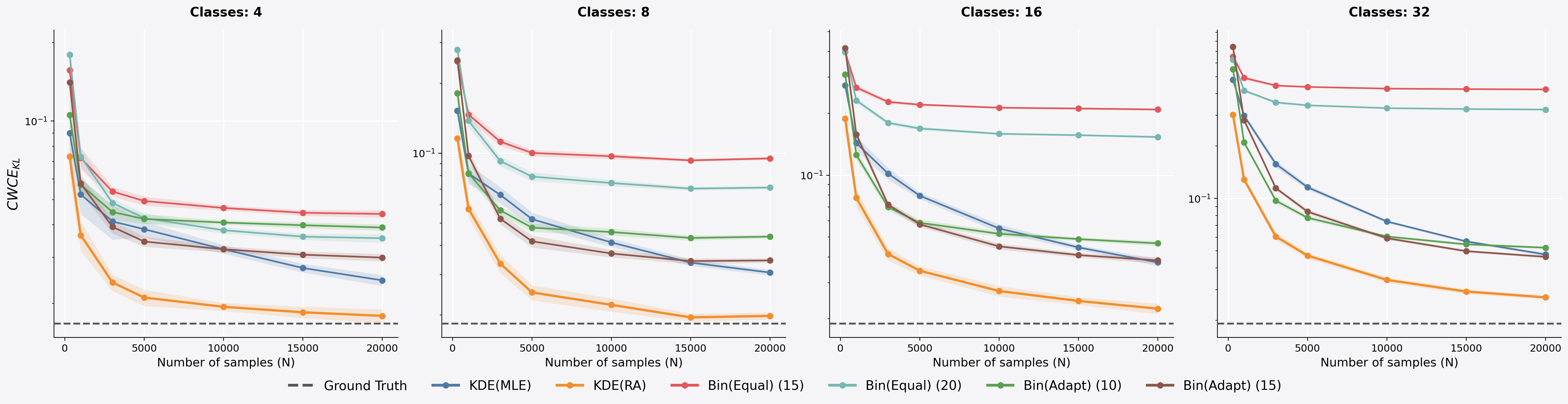}
\caption{(\textbf{Synthetic data} ($t_1=1.0, t_2=0.8$)) Finite sample estimator of CWCE$_{\text{KL}}$: mean and 95\% confidence interval computation across all subsamples of synthetic data.}
\label{fig:syn_classwise_kl}
\end{figure*}

\begin{figure*}[htbp] 
\centering 
\includegraphics[width=1.0\linewidth]{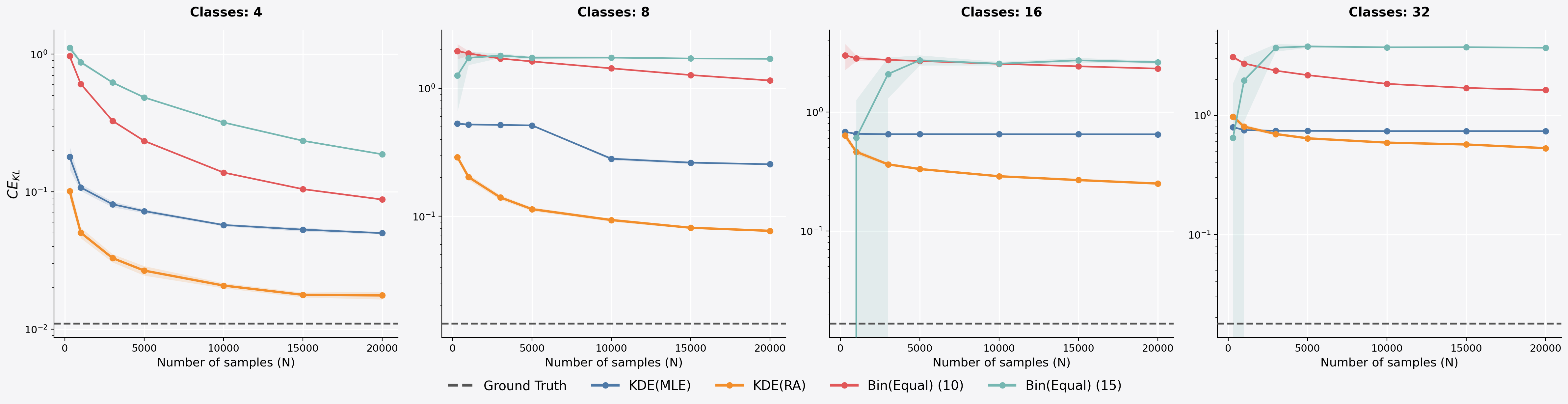}
\caption{(\textbf{Synthetic data} ($t_1=1.0, t_2=0.8$)) Finite sample estimator of CE$_{\text{KL}}^2$: mean and 95\% confidence interval computation across all subsamples of synthetic data.}
\label{fig:syn_canonical_kl}
\end{figure*}

Below we present additional results on the synthetic data, as shown in Table~\ref{tab:synthetic_full_results}-\ref{tab:synthetic_t09t208} and Figs~\ref{fig:syn_classwise_kl}-\ref{fig:syn_canonical_kl}. RA consistently outperforms both discrete binning and standard KDE methods. In the classwise setting, RA achieves the lowest MAE in every scenario, often reducing error by an order of magnitude. This advantage is most pronounced at $K=32$, where RA maintains stable estimates despite the data sparsity that compromises traditional binning. Similarly, in the canonical case, RA dominates the KL divergence results. While KRR emerges as a strong baseline for canonical $L_2$ error in higher dimensions, RA offers a more versatile framework that generalizes across different proper scoring rules and evaluation metrics.

\begin{table*}[htbp]
\caption{Summary of MAE for Calibration Error estimators $\times 10^{2}$ ($\downarrow$) ($t_1=1.0, t_2=0.8$). Best values are in \textcolor{best}{red}, and second-best in \textcolor{second}{teal}. Results are reported at $N=20,000$. Note that Equal Binning uses $n_{\text{bins}} \in \{10, 15\}$ for canonical and $n_{\text{bins}} \in \{15, 20\}$ for classwise setting.}
\centering
\renewcommand{\arraystretch}{1.2}
\setlength{\tabcolsep}{12pt}
\resizebox{\textwidth}{!}{
\begin{tabular}{@{}cclccccccc@{}}
\toprule
\multirow{2}{*}{\textbf{Mode}} & \multirow{2}{*}{\textbf{Metric}} & \multirow{2}{*}{\textbf{\# Classes}} & \multicolumn{2}{c}{\textbf{Equal Binning}} & \multicolumn{2}{c}{\textbf{Adaptive Binning}} & \multicolumn{3}{c}{\textbf{KDE Methods}} \\
\cmidrule(lr){4-5} \cmidrule(lr){6-7} \cmidrule(lr){8-10}
& & & \textbf{$10/15$} & \textbf{$15/20$} & \textbf{10} & \textbf{15} & \textbf{MLE} & \textbf{KRR} & \textbf{RA} \\
\midrule
\multirow{8}{*}{\rotatebox[origin=c]{90}{\textbf{Canonical}}} 
& \multirow{4}{*}{\textbf{CE}$_{\text{KL}}$} 
& 4  & 7.677 & 17.613 & - & - & \textcolor{second}{3.903} & - & \textcolor{best}{0.663} \\
& & 8  & 113.732 & 168.737 & - & - & \textcolor{second}{24.049} & - & \textcolor{best}{6.242} \\
& & 16 & 229.377 & 260.033 & - & - & \textcolor{second}{63.142} & - & \textcolor{best}{23.321} \\
& & 32 & 161.158 & 367.625 & - & - & \textcolor{second}{71.926} & - & \textcolor{best}{51.287} \\
\cmidrule(lr){2-10}
& \multirow{4}{*}{\textbf{CE}$_2^2$} 
& 4  & 1.970 & 6.488 & - & - & 1.654 & \textcolor{second}{0.132} & \textcolor{best}{0.049} \\
& & 8  & 34.362 & 70.000 & - & - & 7.401 & \textcolor{second}{0.462} & \textcolor{best}{0.620} \\
& & 16 & 57.777 & 88.249 & - & - & 6.707 & \textcolor{best}{1.005} & \textcolor{second}{2.047} \\
& & 32 & 18.749 & 92.894 & - & - & 3.968 & \textcolor{best}{0.747} & \textcolor{second}{3.320} \\
\midrule
\multirow{8}{*}{\rotatebox[origin=c]{90}{\textbf{Classwise}}} 
& \multirow{4}{*}{\textbf{CWCE}$_{\text{KL}}$} 
& 4  & 2.732 & 1.882 & 2.235 & 1.323 & \textcolor{second}{0.778} & - & \textcolor{best}{0.176} \\
& & 8  & 7.658 & 5.271 & 2.528 & 1.606 & \textcolor{second}{1.216} & - & \textcolor{best}{0.157} \\
& & 16 & 18.982 & 13.443 & 2.762 & 1.951 & \textcolor{second}{1.868} & - & \textcolor{best}{0.349} \\
& & 32 & 40.411 & 30.510 & 3.306 & \textcolor{second}{2.713} & 2.858 & - & \textcolor{best}{0.794} \\
\cmidrule(lr){2-10}
& \multirow{4}{*}{\textbf{CWCE}$_2^2$} 
& 4  & 0.374 & \textcolor{second}{0.298} & 1.085 & 0.587 & 0.473 & - & \textcolor{best}{0.102} \\
& & 8  & 0.745 & \textcolor{second}{0.558} & 1.794 & 1.116 & 0.769 & - & \textcolor{best}{0.205} \\
& & 16 & 1.383 & \textcolor{second}{1.003} & 1.653 & 1.127 & 1.017 & - & \textcolor{best}{0.341} \\
& & 32 & 2.158 & 1.473 & 1.125 & \textcolor{second}{0.818} & 1.008 & - & \textcolor{best}{0.379} \\
\bottomrule
\end{tabular}}
\label{tab:synthetic_full_results}
\end{table*}

\begin{table*}[htbp]
\caption{Summary of MAE for Calibration Error estimators $\times 10^{2}$ ($\downarrow$) ($t_1=0.9, t_2=0.6$). Best values are in \textcolor{best}{red}, and second-best in \textcolor{second}{teal}. Results are reported at $N=20,000$. Note that Equal Binning uses $n_{\text{bins}} \in \{10, 15\}$ for canonical and $n_{\text{bins}} \in \{15, 20\}$ for classwise setting.}
\centering
\renewcommand{\arraystretch}{1.2}
\setlength{\tabcolsep}{12pt}
\resizebox{\textwidth}{!}{
\begin{tabular}{@{}cclccccccc@{}}
\toprule
\multirow{2}{*}{\textbf{Mode}} & \multirow{2}{*}{\textbf{Metric}} & \multirow{2}{*}{\textbf{\# Classes}} & \multicolumn{2}{c}{\textbf{Equal Binning}} & \multicolumn{2}{c}{\textbf{Adaptive Binning}} & \multicolumn{3}{c}{\textbf{KDE Methods}} \\
\cmidrule(lr){4-5} \cmidrule(lr){6-7} \cmidrule(lr){8-10}
& & & \textbf{$10/15$} & \textbf{$15/20$} & \textbf{10} & \textbf{15} & \textbf{MLE} & \textbf{KRR} & \textbf{RA} \\
\midrule
\multirow{8}{*}{\rotatebox[origin=c]{90}{\textbf{Canonical}}} 
& \multirow{4}{*}{\textbf{CE}$_{\text{KL}}$} 
& 4  & 14.558 & 21.008 & - & - & \textcolor{second}{11.575} & - & \textcolor{best}{0.791} \\
& & 8  & 109.049 & 143.685 & - & - & \textcolor{second}{51.455} & - & \textcolor{best}{9.287} \\
& & 16 & 249.979 & 270.066 & - & - & \textcolor{second}{114.399} & - & \textcolor{best}{35.361} \\
& & 32 & 329.954 & 391.260 & - & - & \textcolor{second}{117.565} & - & \textcolor{best}{79.370} \\
\cmidrule(lr){2-10}
& \multirow{4}{*}{\textbf{CE}$_2^2$} 
& 4  & 2.094 & 6.127 & - & - & 5.381 & \textcolor{best}{0.140} & \textcolor{second}{0.406} \\
& & 8  & 28.053 & 48.029 & - & - & 17.594 & \textcolor{second}{0.464} & \textcolor{best}{0.049} \\
& & 16 & 60.181 & 76.850 & - & - & 26.899 & \textcolor{best}{0.967} & \textcolor{second}{1.379} \\
& & 32 & 52.826 & 93.387 & - & - & 6.776 & \textcolor{best}{1.462} & \textcolor{second}{3.313} \\
\midrule
\multirow{8}{*}{\rotatebox[origin=c]{90}{\textbf{Classwise}}} 
& \multirow{4}{*}{\textbf{CWCE}$_{\text{KL}}$} 
& 4  & 8.670 & 6.533 & 3.720 & \textcolor{second}{2.117} & 0.764 & - & \textcolor{best}{0.492} \\
& & 8  & 18.642 & 14.312 & 4.232 & \textcolor{second}{2.483} & 1.069 & - & \textcolor{best}{0.991} \\
& & 16 & 37.257 & 29.019 & 4.694 & \textcolor{second}{2.893} & 1.896 & - & \textcolor{best}{1.085} \\
& & 32 & 65.637 & 53.519 & 5.544 & \textcolor{second}{3.919} & 3.322 & - & \textcolor{best}{0.717} \\
\cmidrule(lr){2-10}
& \multirow{4}{*}{\textbf{CWCE}$_2^2$} 
& 4  & 0.194 & \textcolor{second}{0.177} & 0.989 & 0.414 & 0.449 & - & \textcolor{best}{0.227} \\
& & 8  & 0.467 & \textcolor{second}{0.316} & 2.783 & 1.538 & 0.877 & - & \textcolor{best}{0.166} \\
& & 16 & 1.230 & \textcolor{second}{0.944} & 3.569 & 2.262 & 1.475 & - & \textcolor{best}{0.236} \\
& & 32 & 2.037 & \textcolor{second}{1.505} & 2.865 & 1.994 & 1.641 & - & \textcolor{best}{0.374} \\
\bottomrule
\end{tabular}}
\label{tab:synthetic_results_t09_t06}
\end{table*}

\begin{table*}[htbp]
\caption{Summary of MAE for Calibration Error estimators $\times 10^{2}$ ($\downarrow$) ($t_1=0.9, t_2=0.7$). Best values are in \textcolor{best}{red}, and second-best in \textcolor{second}{teal}. Results are reported at $N=20,000$. Note that Equal Binning uses $n_{\text{bins}} \in \{10, 15\}$ for canonical and $n_{\text{bins}} \in \{15, 20\}$ for classwise setting.}
\centering
\renewcommand{\arraystretch}{1.2}
\setlength{\tabcolsep}{12pt}
\resizebox{\textwidth}{!}{
\begin{tabular}{@{}cclccccccc@{}}
\toprule
\multirow{2}{*}{\textbf{Mode}} & \multirow{2}{*}{\textbf{Metric}} & \multirow{2}{*}{\textbf{\# Classes}} & \multicolumn{2}{c}{\textbf{Equal Binning}} & \multicolumn{2}{c}{\textbf{Adaptive Binning}} & \multicolumn{3}{c}{\textbf{KDE Methods}} \\
\cmidrule(lr){4-5} \cmidrule(lr){6-7} \cmidrule(lr){8-10}
& & & \textbf{$10/15$} & \textbf{$15/20$} & \textbf{10} & \textbf{15} & \textbf{MLE} & \textbf{KRR} & \textbf{RA} \\
\midrule
\multirow{8}{*}{\rotatebox[origin=c]{90}{\textbf{Canonical}}} 
& \multirow{4}{*}{\textbf{CE}$_{\text{KL}}$} 
& 4  & 10.656 & 19.042 & - & - & \textcolor{second}{8.052} & - & \textcolor{best}{0.250} \\
& & 8  & 110.227 & 151.600 & - & - & \textcolor{second}{38.832} & - & \textcolor{best}{7.510} \\
& & 16 & 246.570 & 260.973 & - & - & \textcolor{second}{86.389} & - & \textcolor{best}{28.865} \\
& & 32 & 266.295 & 384.407 & - & - & \textcolor{second}{98.759} & - & \textcolor{best}{66.036} \\
\cmidrule(lr){2-10}
& \multirow{4}{*}{\textbf{CE}$_2^2$} 
& 4  & 1.997 & 6.406 & - & - & 3.587 & \textcolor{best}{0.133} & \textcolor{second}{0.234} \\
& & 8  & 31.402 & 56.564 & - & - & 12.783 & \textcolor{second}{0.490} & \textcolor{best}{0.219} \\
& & 16 & 63.588 & 82.642 & - & - & 19.480 & \textcolor{best}{1.018} & \textcolor{second}{1.674} \\
& & 32 & 37.539 & 94.876 & - & - & 5.718 & \textcolor{best}{1.155} & \textcolor{second}{3.273} \\
\midrule
\multirow{8}{*}{\rotatebox[origin=c]{90}{\textbf{Classwise}}} 
& \multirow{4}{*}{\textbf{CWCE}$_{\text{KL}}$} 
& 4  & 5.297 & 3.866 & 3.034 & 1.757 & \textcolor{second}{0.816} & - & \textcolor{best}{0.237} \\
& & 8  & 12.734 & 9.250 & 3.530 & 2.145 & \textcolor{second}{1.243} & - & \textcolor{best}{0.307} \\
& & 16 & 27.723 & 20.684 & 3.935 & 2.546 & \textcolor{second}{1.996} & - & \textcolor{best}{0.329} \\
& & 32 & 53.301 & 41.988 & 4.640 & 3.438 & \textcolor{second}{3.304} & - & \textcolor{best}{0.243} \\
\cmidrule(lr){2-10}
& \multirow{4}{*}{\textbf{CWCE}$_2^2$} 
& 4  & 0.297 & \textcolor{second}{0.242} & 1.104 & 0.538 & 0.484 & - & \textcolor{best}{0.100} \\
& & 8  & 0.656 & \textcolor{second}{0.500} & 2.503 & 1.461 & 0.865 & - & \textcolor{best}{0.099} \\
& & 16 & 1.344 & \textcolor{second}{1.003} & 2.811 & 1.834 & 1.290 & - & \textcolor{best}{0.314} \\
& & 32 & 2.153 & 1.546 & 2.121 & 1.494 & \textcolor{second}{1.424} & - & \textcolor{best}{0.439} \\
\bottomrule
\end{tabular}}
\label{tab:synthetic_full_results_t09_t07}
\end{table*}

\begin{table*}[htbp]
\caption{Summary of MAE for Calibration Error estimators $\times 10^{2}$ ($\downarrow$) ($t_1=0.9, t_2=0.8$). Best values are in \textcolor{best}{red}, and second-best in \textcolor{second}{teal}. Results are reported at $N=20,000$. Note that Equal Binning uses $n_{\text{bins}} \in \{10, 15\}$ for canonical and $n_{\text{bins}} \in \{15, 20\}$ for classwise setting.}
\centering
\renewcommand{\arraystretch}{1.2}
\setlength{\tabcolsep}{12pt}
\resizebox{\textwidth}{!}{
\begin{tabular}{@{}cclccccccc@{}}
\toprule
\multirow{2}{*}{\textbf{Mode}} & \multirow{2}{*}{\textbf{Metric}} & \multirow{2}{*}{\textbf{\# Classes}} & \multicolumn{2}{c}{\textbf{Equal Binning}} & \multicolumn{2}{c}{\textbf{Adaptive Binning}} & \multicolumn{3}{c}{\textbf{KDE Methods}} \\
\cmidrule(lr){4-5} \cmidrule(lr){6-7} \cmidrule(lr){8-10}
& & & \textbf{$10/15$} & \textbf{$15/20$} & \textbf{10} & \textbf{15} & \textbf{MLE} & \textbf{KRR} & \textbf{RA} \\
\midrule
\multirow{8}{*}{\rotatebox[origin=c]{90}{\textbf{Canonical}}} 
& \multirow{4}{*}{\textbf{CE}$_{\text{KL}}$} 
& 4  & 8.636 & 17.911 & - & - & \textcolor{second}{5.501} & - & \textcolor{best}{0.511} \\
& & 8  & 110.441 & 157.466 & - & - & \textcolor{second}{29.843} & - & \textcolor{best}{6.237} \\
& & 16 & 236.318 & 255.321 & - & - & \textcolor{second}{71.770} & - & \textcolor{best}{24.707} \\
& & 32 & 205.838 & 371.927 & - & - & \textcolor{second}{83.543} & - & \textcolor{best}{56.561} \\
\cmidrule(lr){2-10}
& \multirow{4}{*}{\textbf{CE}$_2^2$} 
& 4  & 1.996 & 6.459 & - & - & 2.367 & \textcolor{second}{0.137} & \textcolor{best}{0.085} \\
& & 8  & 33.175 & 63.510 & - & - & 9.604 & \textcolor{second}{0.483} & \textcolor{best}{0.444} \\
& & 16 & 61.803 & 84.961 & - & - & 7.814 & \textcolor{best}{1.026} & \textcolor{second}{1.889} \\
& & 32 & 26.573 & 94.227 & - & - & 4.789 & \textcolor{best}{0.922} & \textcolor{second}{3.324} \\
\midrule
\multirow{8}{*}{\rotatebox[origin=c]{90}{\textbf{Classwise}}} 
& \multirow{4}{*}{\textbf{CWCE}$_{\text{KL}}$} 
& 4  & 3.428 & 2.416 & 2.504 & \textcolor{second}{1.467} & 0.817 & - & \textcolor{best}{0.166} \\
& & 8  & 8.973 & 6.299 & 2.959 & \textcolor{second}{1.852} & 1.283 & - & \textcolor{best}{0.148} \\
& & 16 & 21.506 & 15.449 & 3.344 & \textcolor{second}{2.261} & 1.988 & - & \textcolor{best}{0.295} \\
& & 32 & 44.723 & 34.119 & 3.918 & \textcolor{second}{3.056} & 3.108 & - & \textcolor{best}{0.669} \\
\cmidrule(lr){2-10}
& \multirow{4}{*}{\textbf{CWCE}$_2^2$} 
& 4  & 0.400 & \textcolor{second}{0.304} & 1.144 & 0.602 & 0.491 & - & \textcolor{best}{0.092} \\
& & 8  & 0.745 & \textcolor{second}{0.567} & 2.167 & 1.319 & 0.831 & - & \textcolor{best}{0.202} \\
& & 16 & 1.397 & \textcolor{second}{1.031} & 2.197 & 1.469 & 1.153 & - & \textcolor{best}{0.356} \\
& & 32 & 2.213 & 1.554 & 1.577 & \textcolor{second}{1.130} & 1.243 & - & \textcolor{best}{0.427} \\
\bottomrule
\end{tabular}}
\label{tab:synthetic_t09t208}
\end{table*}

\subsection{Real-world data} 
\begin{figure*}[htbp] 
\centering 
\includegraphics[width=1.0\linewidth]{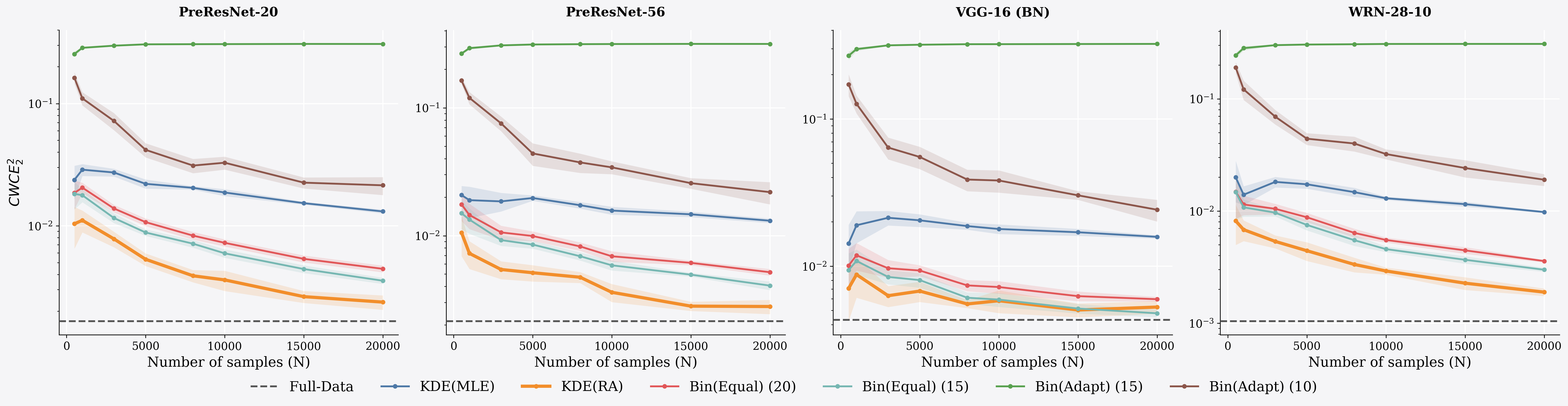}
\caption{(\textbf{CIFAR10}) Finite sample estimator of CWCE$_2^2$: mean and 95\% confidence interval computation across all subsamples.}
\label{fig:cifar10_classwise_l2}
\end{figure*}

\begin{figure*}[htbp] 
\centering 
\includegraphics[width=1.0\linewidth]{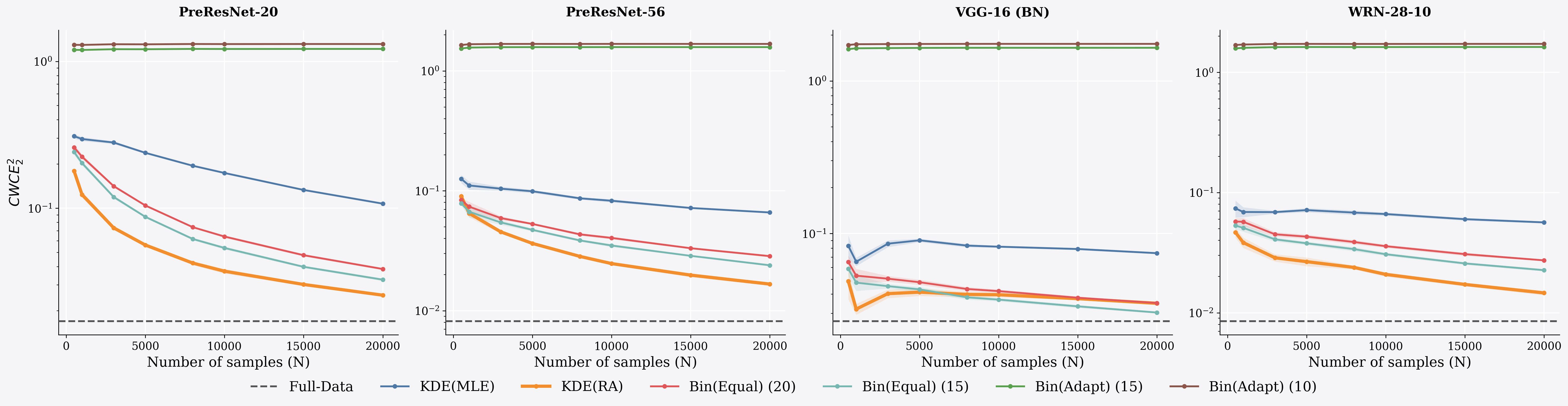}
\caption{(\textbf{CIFAR100}) Finite sample estimator of CWCE$_2^2$: mean and 95\% confidence interval computation across all subsamples.}
\label{fig:cifar100_classwise_l2}
\end{figure*}

\begin{figure*}[htbp] 
\centering 
\includegraphics[width=1.0\linewidth]{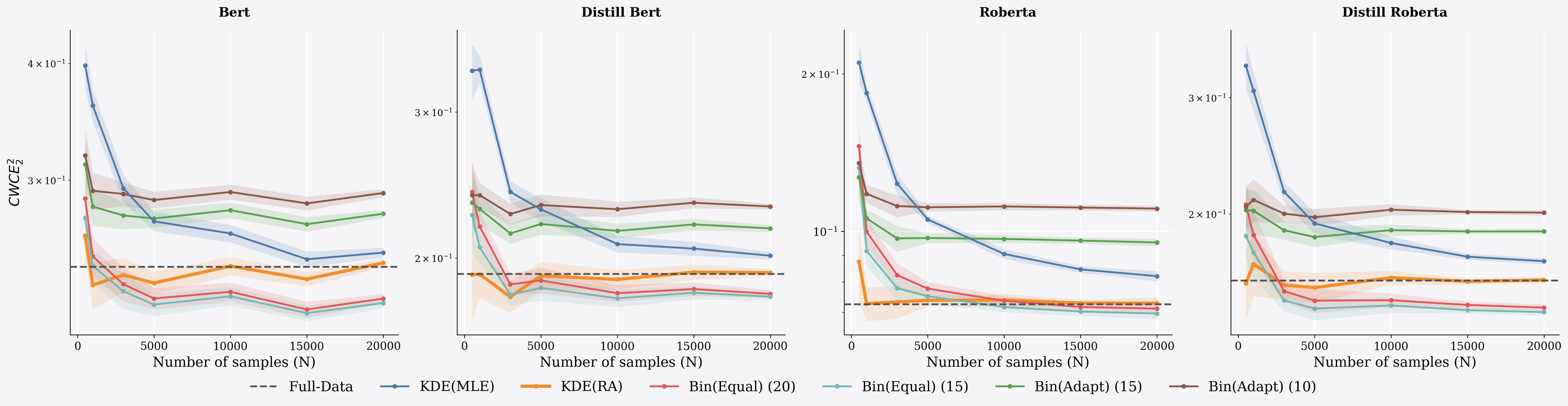}
\caption{(\textbf{Amazon}) Finite sample estimator of CWCE$_2^2$: mean and 95\% confidence interval computation across all subsamples.}
\label{fig:Amazon_classwise_l2}
\end{figure*}

\begin{figure*}[htbp] 
\centering 
\includegraphics[width=1.0\linewidth]{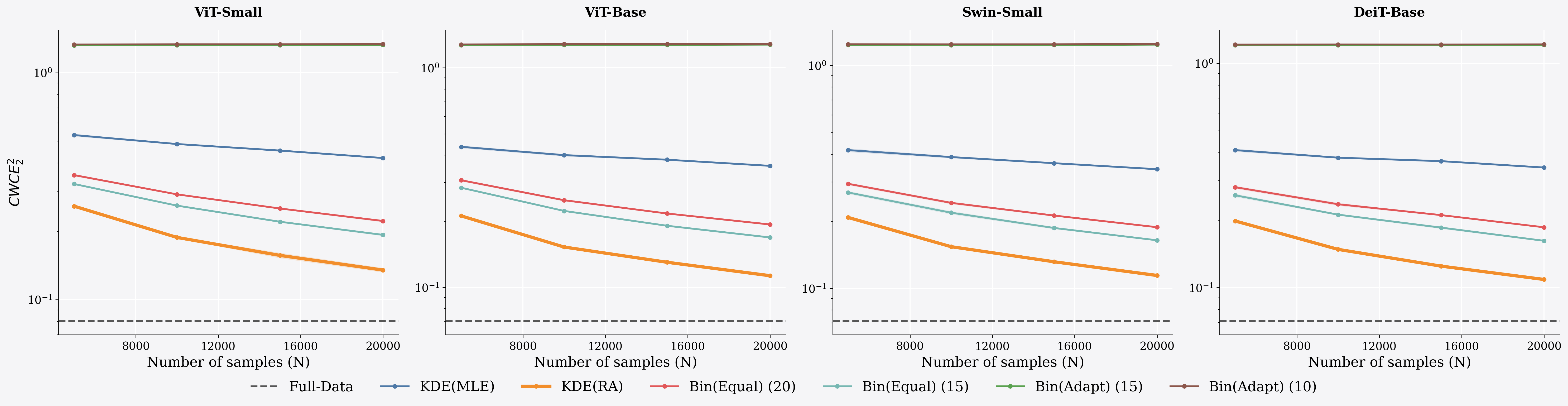}
\caption{(\textbf{ImageNet}) Finite sample estimator of CWCE$_2^2$: mean and 95\% confidence interval computation across all subsamples.}
\label{fig:image_classwise_l2}
\end{figure*}

\begin{figure*}[htbp] 
\centering 
\includegraphics[width=1.0\linewidth]{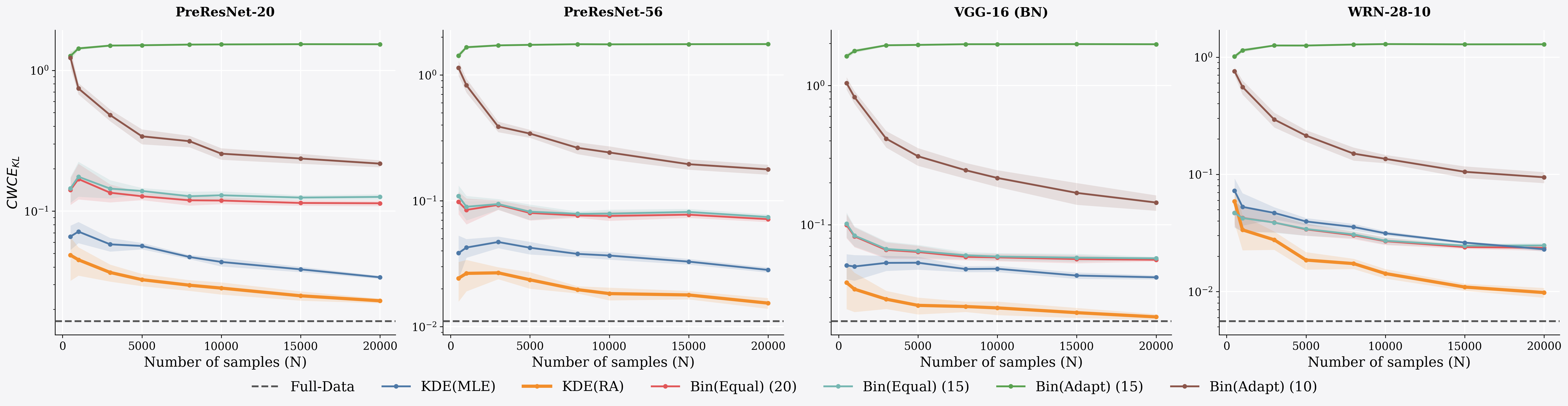}
\caption{(\textbf{CIFAR10}) Finite sample estimator of CWCE$_{\text{KL}}^2$: mean and 95\% confidence interval computation across all subsamples.}
\label{fig:cifar10_classwise_kl}
\end{figure*}

\begin{figure*}[htbp] 
\centering 
\includegraphics[width=1.0\linewidth]{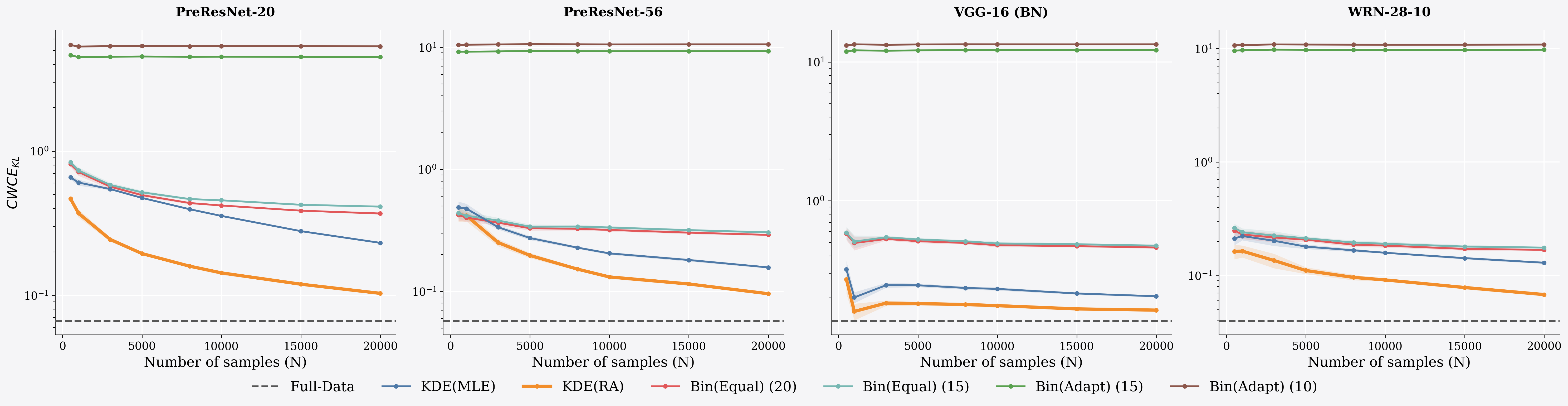}
\caption{(\textbf{CIFAR100}) Finite sample estimator of CWCE$_{\text{KL}}^2$: mean and 95\% confidence interval computation across all subsamples.}
\label{fig:cifar100_classwise_kl}
\end{figure*}

\begin{figure*}[htbp] 
\centering 
\includegraphics[width=1.0\linewidth]{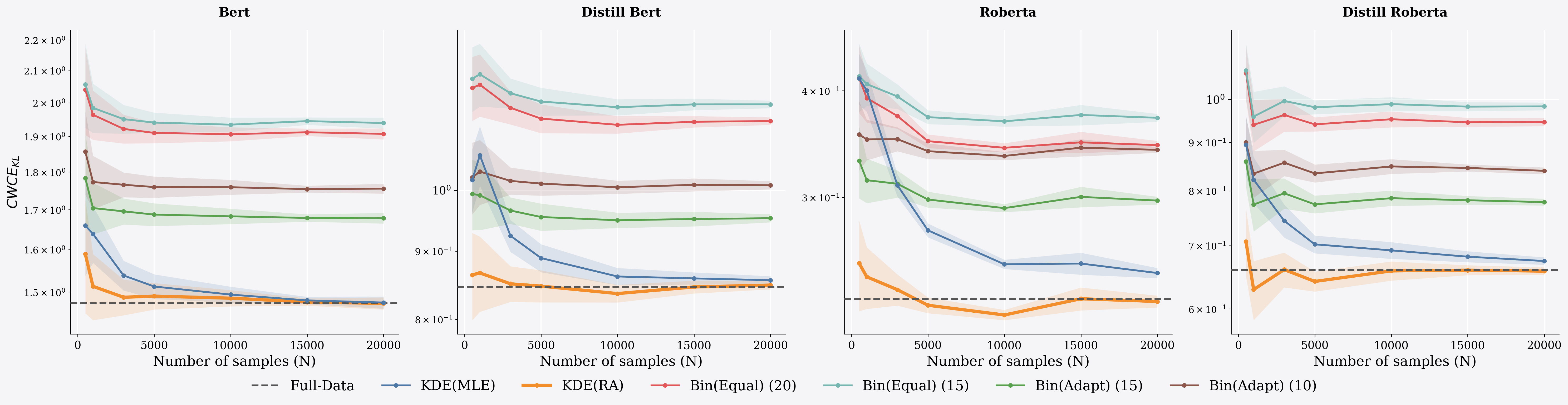}
\caption{(\textbf{Amazon}) Finite sample estimator of CWCE$_{\text{KL}}^2$: mean and 95\% confidence interval computation across all subsamples.}
\label{fig:Amazon_classwise_kl}
\end{figure*}

\begin{figure*}[htbp] 
\centering 
\includegraphics[width=1.0\linewidth]{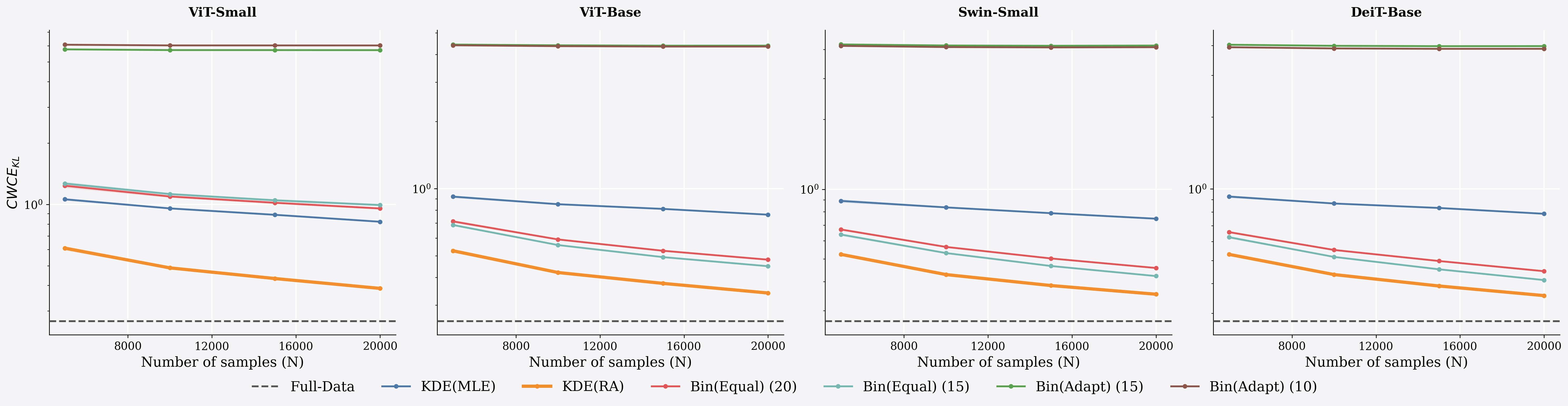}
\caption{(\textbf{ImageNet}) Finite sample estimator of CWCE$_{\text{KL}}^2$: mean and 95\% confidence interval computation across all subsamples.}
\label{fig:image_classwise_kl}
\end{figure*}

Below we present the result of CWCE$_2^2$, CWCE$_{\text{KL}}^2$, CE$_{\text{KL}}$ and CE$_{2}^2$ estimators across different datasets and architectures. Consistent with our synthetic findings, RA demonstrates clear superiority over the MLE-based approach and traditional binning schemes. Across benchmarks ranging from ResNets to Vision Transformers, RA aligns more closely with the ground truth computed over the full population than those methods, particularly in class-wise settings, as shown in Figs.~\ref{fig:cifar10_classwise_l2}-\ref{fig:image_classwise_kl}. We further observe that KDE performance degrades in canonical settings with many classes, such as CIFAR-100, due to data sparsity on the high-dimensional simplex. In these regimes, KRR remains a strong baseline.

\begin{figure*}[htbp] 
\centering 
\includegraphics[width=1.0\linewidth]{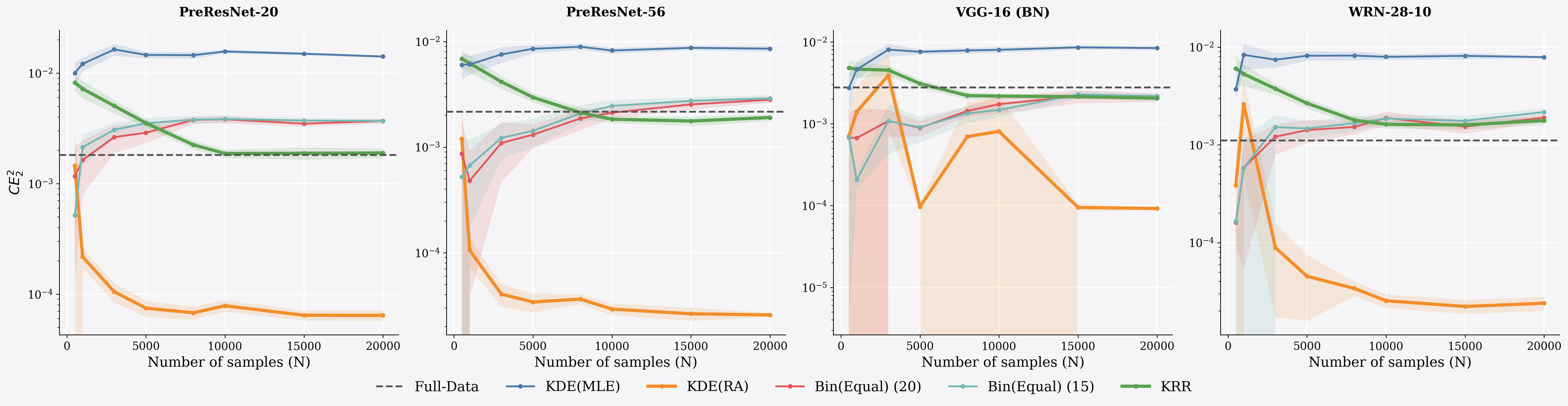}
\caption{(\textbf{CIFAR10}) Finite sample estimator of CE$_{2}^2$: mean and 95\% confidence interval computation across all subsamples.}
\label{fig:cifar10_canonical_l2}
\end{figure*}

\begin{figure*}[htbp] 
\centering 
\includegraphics[width=1.0\linewidth]{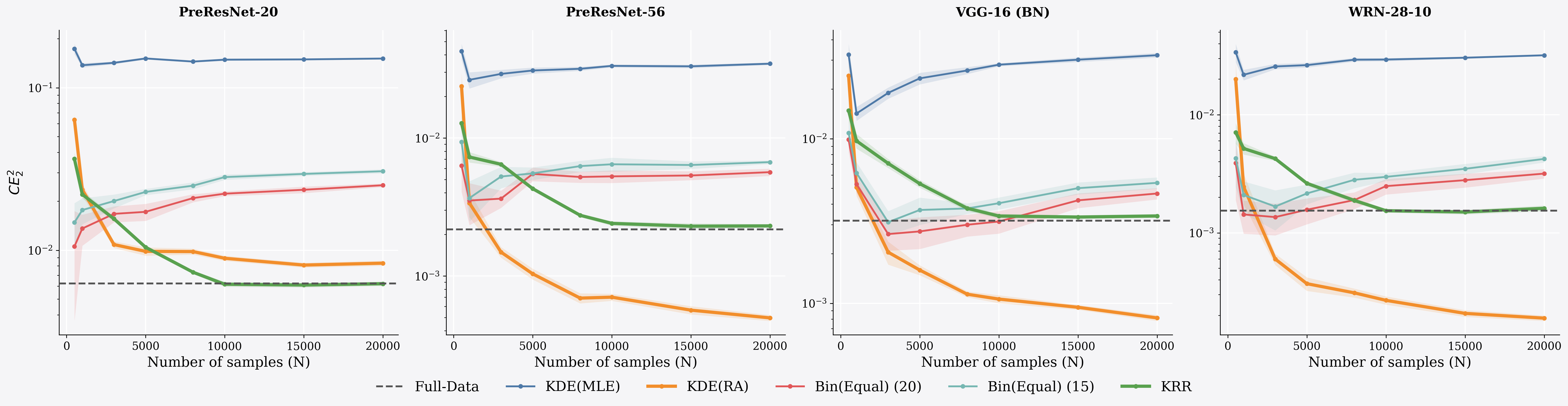}
\caption{(\textbf{CIFAR100}) Finite sample estimator of CE$_{2}^2$: mean and 95\% confidence interval computation across all subsamples.}
\label{fig:cifar100_canonical_l2}
\end{figure*}

\begin{figure*}[htbp] 
\centering 
\includegraphics[width=1.0\linewidth]{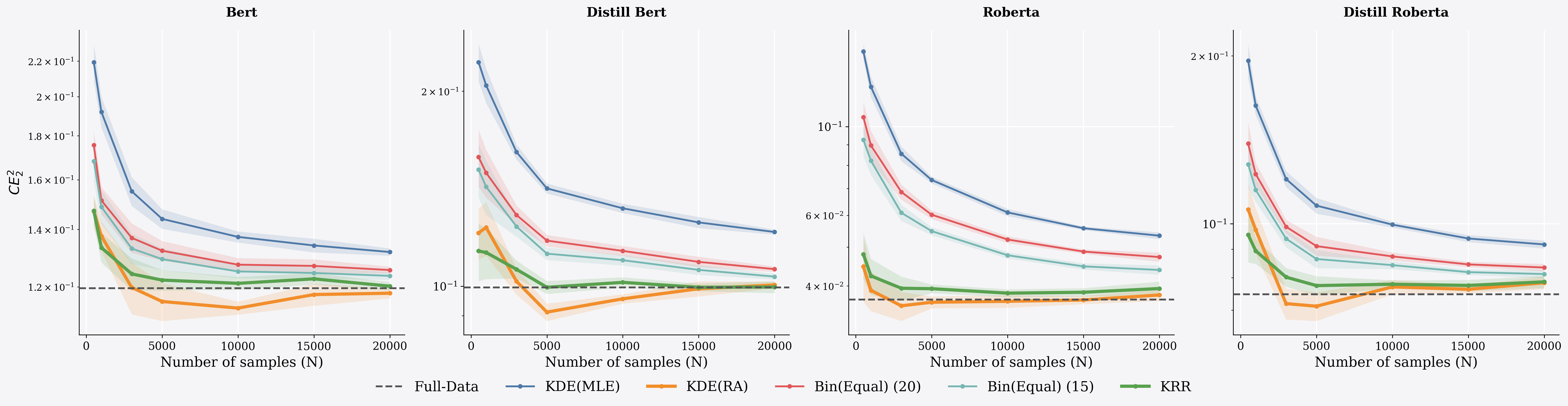}
\caption{(\textbf{Amazon}) Finite sample estimator of CE$_{2}^2$: mean and 95\% confidence interval computation across all subsamples.}
\label{fig:Amazon_canonical_l2}
\end{figure*}

\begin{figure*}[htbp] 
\centering 
\includegraphics[width=1.0\linewidth]{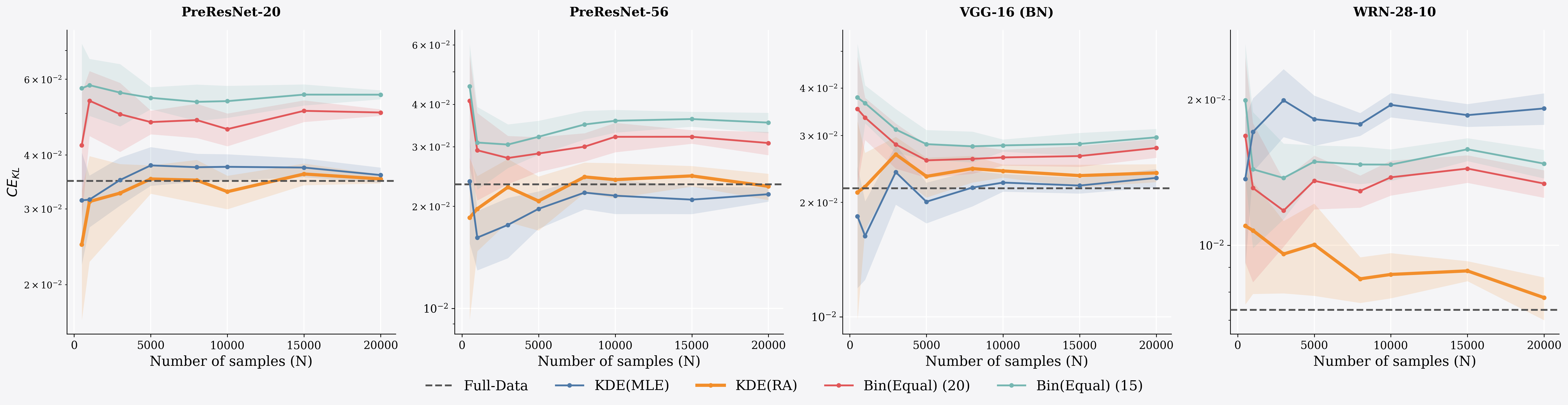}
\caption{(\textbf{CIFAR10}) Finite sample estimator of CE$_{\text{KL}}^2$: mean and 95\% confidence interval computation across all subsamples.}
\label{fig:cifar10_canonical_kl}
\end{figure*}

\begin{figure*}[htbp] 
\centering 
\includegraphics[width=1.0\linewidth]{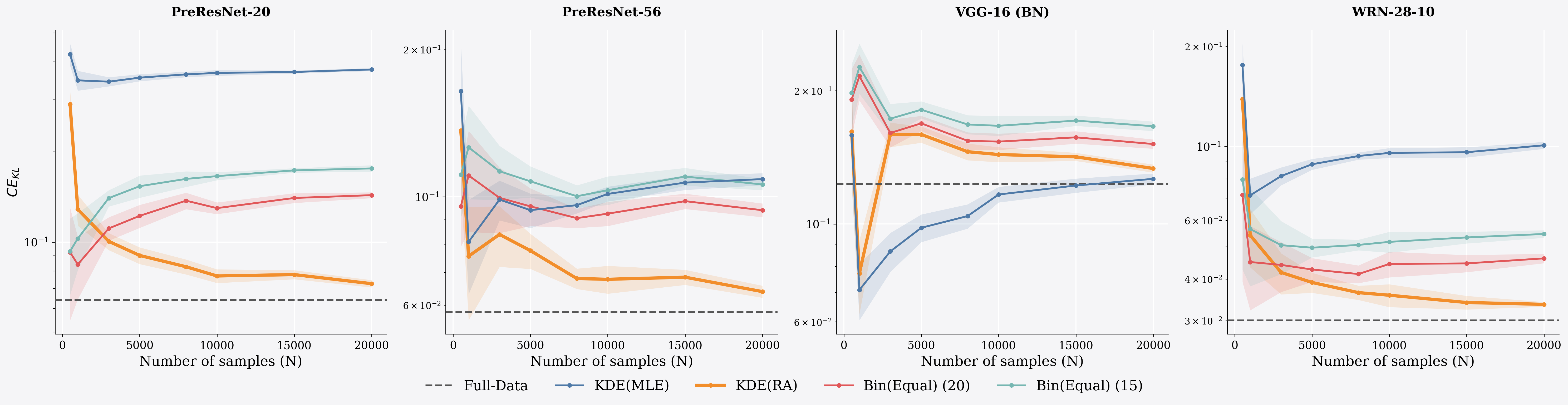}
\caption{(\textbf{CIFAR100}) Finite sample estimator of CE$_{\text{KL}}^2$: mean and 95\% confidence interval computation across all subsamples.}
\label{fig:cifar100_canonical_kl}
\end{figure*}

\begin{figure*}[htbp] 
\centering 
\includegraphics[width=1.0\linewidth]{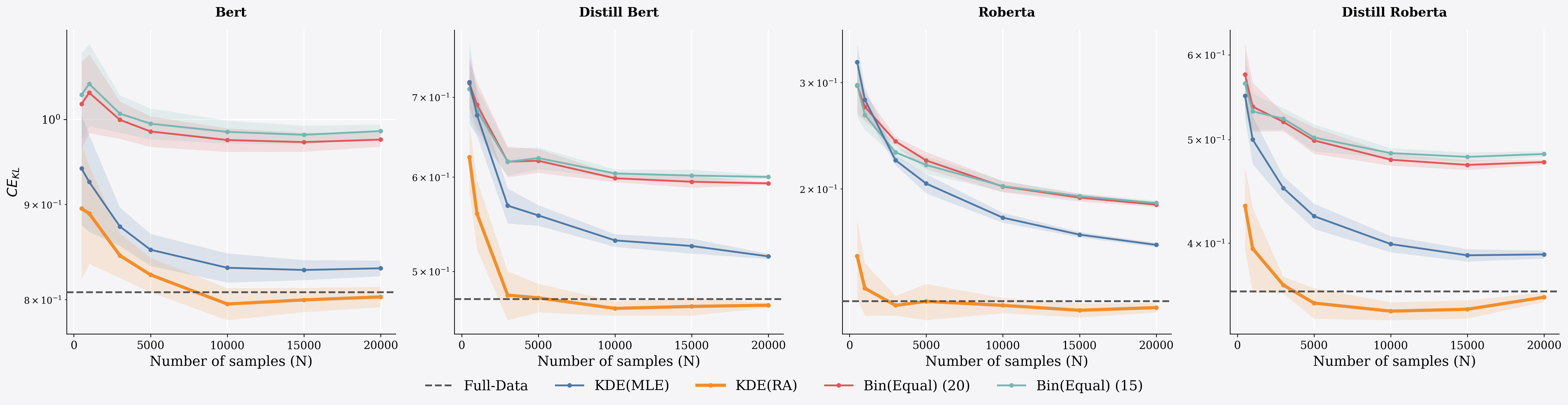}
\caption{(\textbf{Amazon}) Finite sample estimator of CE$_{\text{KL}}^2$: mean and 95\% confidence interval computation across all subsamples.}
\label{fig:Amazon_canonical_kl}
\end{figure*}

\end{document}